%% file: ms.tex
\documentclass{article}

\usepackage{arxiv}

\usepackage[utf8]{inputenc} 
\usepackage[T1]{fontenc}    
\usepackage{hyperref}       
\usepackage{url}            
\usepackage{amsfonts}       
\usepackage{nicefrac}       
\usepackage{microtype}      

\usepackage{graphicx}
\usepackage{natbib}
\usepackage{doi}

\usepackage{amssymb}
\usepackage{amsmath}
\DeclareMathOperator*{\argmax}{arg\,max}

\usepackage{bm}
\usepackage{enumitem}
\usepackage{float}
\usepackage{comment}
\usepackage{graphicx}
\usepackage[export]{adjustbox}

\usepackage{tikz}
\usetikzlibrary{calc}
\usepackage{ifthen}

\usepackage{hyperref}

\usepackage{caption} 
\usepackage[section]{placeins}

\usepackage[acronym]{glossaries}

\newacronym{pls}{PLS}{Partial Least Squares}
\newacronym{cca}{CCA}{Canonical Correlation Analysis}
\newacronym{pca}{PCA}{Principal Component Analysis}
\newacronym{ols}{OLS}{Ordinary Least Squares}
\newacronym{rr}{RR}{Ridge Regression}
\newacronym{en}{EN}{Elastic Net}
\newacronym{lasso}{lasso}{Least Absolute Shrinkage and Selection Operator}
\newacronym{lavade}{LAVADE}{Latent Variable Demonstrator}
\newacronym{lfp}{LFP}{Lithium Iron Phosphate}
\newacronym{pcr}{PCR}{Principal Component Regression}
\newacronym{snr}{SNR}{Signal-to-Noise Ratio}
\newacronym{gui}{GUI}{Graphical User Interface}
\newacronym{rss}{RSS}{Residual Sum of Squares}
\newacronym{rmse}{RMSE}{Root-Mean-Square Error}

\title{Latent Variable Method Demonstrator -- Software for Understanding Multivariate Data Analytics Algorithms}

\date{May 15, 2022}

\author{\href{https://orcid.org/0000-0001-8767-4101}{\includegraphics[scale=0.06]{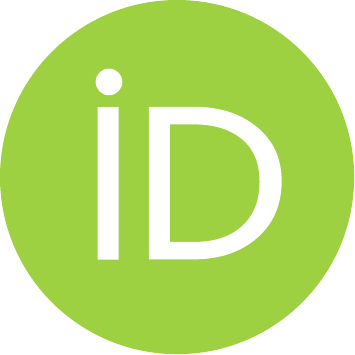}\hspace{1mm} Joachim Schaeffer}\thanks{Parts of this work were done at ETH Zurich.} \\
    Control and Cyber-Physical Systems Laboratory \\
    Technical University of Darmstadt \\ 
    Karolinenpl. 5 \\
    Darmstadt, 64289, Germany
\And
\href{https://orcid.org/
0000-0003-4304-3484}{\includegraphics[scale=0.06]{orcid.pdf}\hspace{1mm}  Richard D. Braatz} \\ 
Massachusetts Institute of Technology \\ 
77 Massachusetts Avenue \\ 
Cambridge, 02139, MA, USA \\
\texttt{braatz@mit.edu}
}
\renewcommand{\headeright}{}
\renewcommand{\undertitle}{}
\renewcommand{\shorttitle}{LAVADE}

\hypersetup{
pdftitle={LAVADE - Latent Variable Method Demonstrator},
pdfsubject={stat.ML},
pdfauthor={Joachim Schaeffer, Richard Braatz},
pdfkeywords={Education, Teaching, Data Analytics, Software},
}

\begin{document}
\maketitle

\input{0Abstract}


\input{1Introduction}

\input{2Regression}

\input{3DemonstratorDesign}

\input{4Datasets}

\input{5IllExamples}

\input{6Conclusion}

\input{Aknowledgements}

\bibliographystyle{unsrt_custom} 
\bibliography{bib_final}

\end{document}

%% file: 0Abstract.tex
\begin{abstract}

The ever-increasing quantity of multivariate process data is driving a need for skilled engineers to analyze, interpret, and build models from such data. Multivariate data analytics relies heavily on linear algebra, optimization, and statistics and can be challenging for students to understand given that most curricula do not have strong coverage in the latter three topics. This article describes interactive software -- the \acrfull{lavade} -- for teaching, learning, and understanding latent variable methods. In this software, users can interactively compare latent variable methods such as \acrfull{pls}, and \acrfull{pcr} with other regression methods such as \acrfull{lasso}, \acrfull{rr}, and \acrfull{en}.  \acrshort{lavade} helps to build intuition on choosing appropriate methods, hyperparameter tuning, and model coefficient interpretation, fostering a conceptual understanding of the algorithms' differences. The software contains a data generation method and three chemical process datasets, allowing for comparing results of datasets with different levels of complexity. 
\acrshort{lavade} is released as open-source software so that others can apply and advance the tool for use in teaching or research.
\end{abstract}

%% file: 1Introduction.tex
\section{Introduction}
\label{sec:Introduction}
Latent variable methods such as partial least squares are among the most widely applied data analytics tools for applications to chemical, physical, and biological processes. It can be challenging to teach these multivariable statistical methods in a way that the students are able to consistently apply these methods effectively in practice, especially considering that many undergraduate programs do not require that their students take courses in linear algebra or applied statistics. The result is that many practitioners use alternative methods instead, such as correlating a peak absorbance to each concentration when calibrating a spectral measurement, which can produce models of much lower accuracy. Alternatively, the latent variable methods are often used in a black-box manner, i.e., without understanding when a situation occurs in which the methods should not be applied, or how to revise the way that data are fed to the method to work around some imperfections in the data such as sensor bias. 
After all, many software packages are available for applying latent variable methods, and it might seem that a deep understanding is not needed for problem-solving. 

A lack of understanding, however, makes it challenging to explain and interpret the models created by these methods, particularly in relating the models generated with the chemistry/physics/biology occurring in the process, to reconcile the data analytics results with domain knowledge. The synthesis of domain knowledge with data analytics is the added value of a well-trained applied data scientist/engineer and is likely to result in the best solutions for the particular problem at hand. Furthermore, preliminary black-box results can lead to choosing overly complicated models that overfit the data. Advancing the understanding and intuition on latent variable methods are needed to avoid overfitting for some types of biased data and ultimately assure model interpretability, leading to higher value, acceptance, and applicability. 

This article describes software and examples developed to train students to achieve a deep understanding of latent variable methods \citep[e.g.,][]{chiang2000fault, Mardia2003} and the related machine learning methods of lasso \citep{Tibshirani1996}, ridge regression \citep{Hoerl1970}, and elastic net \citep{Zou2005}. The graphical user interface is designed for the explicit purpose of teaching undergraduate and graduate students, which is a distinguishing feature from the graphical user interfaces in existing chemometrics software packages that are focused on just applying a method to a dataset. The software takes the perspective of the optimization being solved to help the understanding of the relationship between the latent variable method selected and the results produced.

This tool, referred to as the \acrfull{lavade}, compares a wide range of latent variable regression and other techniques for four different datasets. 
The examples are designed to be easy to understand, and various options to customize the problem are available to learn precisely how the different algorithms approach the model construction.

This article is organized as follows. Section \ref{sec:RegressionTechniques} describes the latent variable and other regression techniques in the software. Section \ref{sec:Demonstrator} describes the software design and architecture from a high-level perspective, and Section \ref{sec:datasets} describes the datasets. Section \ref{sec:example} shows illustrative examples of the software and is followed by the conclusion. 

%% file: 2Regression.tex
\section{Regression Techniques}
\label{sec:RegressionTechniques}


The regression models considered in this article are linear, static models with the general form
\begin{equation}
   y= \beta_0x_0 + \beta_1 x_1 + \cdots + \beta_{n-1} x_{n-1} + \epsilon     
\end{equation}
where 
the output $y$, inputs $x_i$, and regression coefficients $\beta_i$ are real scalars;
$x_0 = 1$; $n$ is the number of inputs and regression coefficients; and $\epsilon$ denotes the error term. Without loss of generality, this model can be rewritten in matrix form as
\begin{equation}
\mathbf{y} =
    \mathbf{X}\boldsymbol{\beta} + \boldsymbol{\epsilon} 
    \label{eq:regression}
\end{equation}
where the input data matrix $\mathbf{X} \in \mathbb{R}^{m \times n}$, the vector of regression coefficients $\boldsymbol{\beta} \in \mathbb{R}^{n \times 1}$,  $\mathbf{y} \in \mathbb{R}^{m \times 1}$, and $m$ is the number of observations. The elements of $\mathbf{X}$ can be any mix of raw input data and transformations of the raw data (aka features). The errors $\boldsymbol{\epsilon}$ are assumed to be homoscedastic, have zero mean, and are uncorrelated. 

Model building involves determining the vector $\boldsymbol{\beta}$ from the data $\mathbf{X}$ and $\mathbf{y}$ that minimizes the error $\boldsymbol{\epsilon}$ concerning a defined measure of the error. 
The goal of multivariate regression is to find a good solution for $\boldsymbol{\beta}$ when there is a relatively low number of observations relative to the number of regression coefficients. For some specific applications, such as building a model to relate spectra to concentrations, \eqref{eq:regression} may not even have a unique solution.  

Some special cases of  \eqref{eq:regression} of interest are:
\begin{enumerate}[label=(\alph*)]
    \item The case where the input data matrix $\mathbf{X}$ has more rows than columns ($m > n$) and  has full column rank, i.e., $\text{rank}(\mathbf{X}) = n$. This case arises when many samples are available for building a model for predicting the output $\mathbf{y}$ from a small number of inputs (or calculated features). It is thus very unlikely that any vector $\mathbf{y} \in \mathbb{R}^{m \times 1}$ is in the column space of $\mathbf{X}$, due to the fact that the columns of $\mathbf{X}$ span only a small subspace of $\mathbb{R}^m$.  
    \label{enum:cm>n}
    \item The case where the input data matrix $\mathbf{X}$ has fewer rows than columns ($m < n$) and $\mathbf{X}$ has full row rank, i.e., $\text{rank}(\mathbf{X}) = m$.  This case can arise, for example, when spectral data are available for a large number of frequencies for building a model for predicting concentration from a relatively small number of samples of known concentration. In this case, the vector $\mathbf{y}$ lies in the column space of $\mathbf{X}$, and there exists no unique solution to \eqref{eq:regression}. This case can be interpreted from the perspective of the right nullspace of $\mathbf{X}$, $\boldsymbol{N}(\mathbf{X})$,  which is the set of nonzero vectors $\mathbf{v}$ that satisfy $\mathbf{Xv}=\mathbf{0}$. In this case, this nullspace is not empty, that is, the nullspace consists of more vectors than the zero vector ($\mathbf{0}$). The vectors can be used to define a basis for the nullspace, and subsequently used to parameterize the infinite set of vectors of regression coefficients $\boldsymbol{\beta}$ that satisfy $\mathbf{y} =
    \mathbf{X}\boldsymbol{\beta}$.  
    \label{enum:cm<n}
    \item The case where $m = n$ and $\text{rank}(\mathbf{X}) = m = n$, for which \eqref{eq:regression} has a unique solution for $\mathbf{y} =
    \mathbf{X}\boldsymbol{\beta}$.  
    This case does not occur when working with larger datasets -- due to biophysicochemical constraints such as associated with conservation equations and reaction networks -- and is not considered further in this article. 
\end{enumerate}
Matrices with redundant rows and/or columns can be reduced to cases \ref{enum:cm>n} or \ref{enum:cm<n} by removing redundant rows and/or columns. 

Different approaches find good solutions for cases \ref{enum:cm>n} and \ref{enum:cm<n} by solving
\begin{equation}
    \mathbf{X}_{\textrm{train}}\hspace{2pt}\boldsymbol{\hat{\beta}} = \mathbf{\hat{y}}
    \label{eq:modeltrain}
\end{equation}
where $\boldsymbol{\hat{\beta}}$ is the vector of parameters estimated from the data $\mathbf{X}_{\textrm{train}}$ leading to the response $\mathbf{\hat{y}}$. The latter is related to the measured model outputs by additive noise $\boldsymbol{\epsilon}$.
\begin{equation}
    \mathbf{\hat{y}} + \boldsymbol{\epsilon} = \mathbf{y}.
    \label{eq:error}
\end{equation}
The intent of the model is to be capable of making predictions of unseen data by taking a weighted combination of columns based on the estimated weights $\boldsymbol{\hat{\beta}}$. A fundamental question in statistics is how to find a good (or best) vector of model parameters $\boldsymbol{\hat{\beta}}$.\footnote{Multiple terms are used in the literature to refer to $\boldsymbol{\hat{\beta}}$, including {\em model parameters} and {\em regression coefficients}. From \eqref{eq:modeltrain}, it is clear that   $\boldsymbol{\hat{\beta}}$ can be interpreted as being a vector of weights, that is, the model output is a weighted combination of the model inputs, with the weights being the elements of $\boldsymbol{\hat{\beta}}$.}

This section describes different approaches that have been developed for finding $\boldsymbol{\hat{\beta}}$. To simplify the notation,  $\boldsymbol{\beta}$ and $\mathbf{X}$ are used instead of $\boldsymbol{\hat{\beta}}$ and $\mathbf{X}_{\textrm{train}}$; in all cases, $\boldsymbol{\beta}$ refers to parameters estimated from training data and $\mathbf{X}$ refers to the training data.
The first part of this section is a compact introduction to \acrfull{ols}, \acrfull{lasso}, \acrfull{rr}, and elastic net (EN)  regression. The second part is an introduction to \acrfull{pca}, \acrfull{pcr}, and \acrfull{pls}, which are methods for dimensionality reduction and regression is performed subsequently in the lower dimensional space.

\subsection{Ordinary Least Squares Regression}

The key idea of \acrshort{ols} regression is to find a solution $\boldsymbol{\beta}$ that minimizes the L$_2$-norm of the model error $\boldsymbol{\epsilon}$ \citep{LA_GStrang},
\begin{equation}
    \min_{\boldsymbol{\beta}}  \|\mathbf{y}-\mathbf{X}\boldsymbol{\beta}\|_2^2.
\end{equation}
Practically, this optimization can be solved by multiplying  \eqref{eq:modeltrain} on the left by $\mathbf{X^{\top}}$,
\begin{equation}
    \mathbf{X^{\top}X}\boldsymbol{\beta} = \mathbf{X^{\top}y},
    \label{eq:normal_equation}
\end{equation}
which is known as the normal equation. For Case a, 
$\mathbf{X^{\top}X}$ is invertible, which leads to the analytical solution
\begin{equation}
    \boldsymbol{\beta} = (\mathbf{X^{\top}X})^{-1}\mathbf{X^{\top}y}.
    \label{eq:OLSclosedform}
\end{equation}
This optimization formulation and solution are known as \acrshort{ols}. The Gauss-Markov theorem states that the \acrshort{ols} solution is optimal with the assumed error structure \eqref{eq:error}. 

In Case b,  
$\mathbf{X^{\top}X}$ is singular. A solution for $\boldsymbol{\beta}$ in this case is
\begin{equation}
    \boldsymbol{\beta} = \mathbf{X^{\top}}(\mathbf{XX^{\top})}^{-1}\mathbf{y}
    \label{eq:MinNormSolution}
\end{equation}
which is known as the minimum norm solution \citep{monticelli1999least}.

\subsection{Lasso}

\acrshort{ols} estimates may have low bias but have very large variance for many real-world data analytics problems, resulting in low prediction accuracy on unseen data.
\Acrshort{lasso} is a strategy for addressing this problem by adding the L$_1$-norm of the weights as a penalty to the least-squares objective \citep{Tibshirani1996}, 
\begin{equation}
    \min_{\boldsymbol{\beta}}  \|\mathbf{y}-\mathbf{X}\boldsymbol{\beta}\|_2^2 + \lambda\|\boldsymbol{\beta}\|_1.
    \label{eq:lasso_obj}
\end{equation}
For positive values for $\lambda$, \acrshort{lasso} 
builds a model in which some weights are zero, $\beta_i = 0$, due to the structure of the optimization. The larger the value of $\lambda$, the more weights are forced to zero.

While there exists no closed-form solution for \eqref{eq:lasso_obj}, common solvers are available. The L$_1$-norm penalty on $\mathbf{\beta}$ in \eqref{eq:lasso_obj} can be rewritten as linear constraints and the resulting optimization can be written as a convex quadratic program which can be solved efficiently using convex programming. A visual explanation for how the shape of the constraints and shape of the objective function in the optimization are responsible for the sparse solution can be found in \cite{Tibshirani1996}. 

Models that use only a subset of available columns of the data matrix $\mathbf{X}$ are called {\em sparse models}. \acrshort{lasso} selects a subset of columns to be used in the regression and can potentially lead to increased interpretability of results by removing measurements that are not needed in the model prediction. 

A drawback of \acrshort{lasso} is that, when some columns in $\mathbf{X}$ have a high pairwise correlation, \acrshort{lasso} tends to select only one column and the choice of column can be sensitive to arbitrarily small perturbation in the data \citep{Zou2005}. This randomness reduces the suitability of such models for interpreting the system in terms of which model inputs appear in the model.

\subsection{Ridge Regression}

The underlying motivation for \acrfull{rr} is identical to \acrshort{lasso}. 
The difference between the methods is that \acrfull{rr}  adds a L$_2$-norm penalty to the objective function,
\begin{equation}
    \min_{\boldsymbol{\beta}}  \|\mathbf{y}-\mathbf{X}\boldsymbol{\beta}\|_2^2 + \lambda\|\boldsymbol{\beta}\|_2^2 .
    \label{eq:obj_rr}
\end{equation}
Setting the derivative of the objective function of  \eqref{eq:obj_rr} to zero leads to the closed-form solution \citep{Hoerl1970}
\begin{equation}
    \boldsymbol{\beta} = (\mathbf{X^{\top}X + \lambda \mathbf{I}})^{-1}\mathbf{X^{\top}y}.
    \label{eq:RRclosedform}
\end{equation}
As in \acrshort{lasso}, the L$_2$-norm penalty on $\boldsymbol{\beta}$ in \eqref{eq:obj_rr} can be rewritten as a constraint. 
Due to the shape of the constraint in the resulting optimization, the weights $\beta_i$ in \acrshort{rr} will never reach zero although they can be arbitrarily small \citep{introstatlearning}. Similarly to \acrshort{lasso}, \acrshort{rr} improves the predictive accuracy of the model by introducing a bias that reduces variance in the estimated parameters \citep{Zou2005}. However, models produced by \acrshort{rr} can be challenging to interpret, since all model inputs are retained in the model even for high values of $\lambda$.

\subsection{Elastic Net}

The \acrfull{en} is a combination of \acrshort{lasso} and \acrshort{rr},
\begin{align}
    \min_{\boldsymbol{\beta}} \|\mathbf{y}-\mathbf{X}\boldsymbol{\beta}\|_2^2 + \lambda P(\boldsymbol{\beta})\\
    P(\boldsymbol{\beta}) = \frac{1-\alpha}{2} \|\boldsymbol{\beta}\|_2^2 + \alpha\|\boldsymbol{\beta}\|_1 
    \label{eq:obj_en}
\end{align}
for $\alpha \in (0,1)$ and non-negative values of $\lambda$. The L$_1$-norm provides the ability to 
set unimportant parameters to zero whereas the L$_2$-norm improves robustness in the selection of which parameters to retain in the model. 
\Acrshort{lasso} and \acrshort{rr} are limiting cases of the \acrshort{en}, for $\alpha$ equal to 1 and 0, respectively. 

\subsection{PCA and PCR}

\acrfull{pca} is an approach to manipulating the data matrix $\mathbf{X} \in \mathbb{R}^{m\times n}$. The idea of \acrshort{pca} is to find a lower dimensional representation of $\mathbf{X}$ while conserving the variations in the data.
From an optimization perspective, \acrshort{pca} solves
\begin{equation}
    \max_{\mathbf{\|w\|_2 = 1}} \|\mathbf{Xw}\|_2^2 
    \label{eq:PCA_opt1}
\end{equation}
for each principal component, with subsequent principal components also required to be orthogonal to the previous principal components. \acrshort{pca} finds a weighted combination of the mean subtracted columns of $\mathbf{X}$ retaining maximal variance.  For the first principal component, the constrained optimization \eqref{eq:PCA_opt1} is equivalent to the unconstrained optimization
\begin{equation}
    \mathbf{w_1} = \argmax_\mathbf{W}
    \frac{\mathbf{w}^{\top}\mathbf{X}^{\top}\mathbf{Xw}}{\mathbf{w}^{\top}\mathbf{w}}.
    \label{eq:PCA_opt2}
\end{equation} 
The optimization \eqref{eq:PCA_opt2} is solved by $\mathbf{w}$ being the eigenvector corresponding to the largest eigenvalue of the positive semidefinite matrix $\mathbf{X}^{\top}\mathbf{X}$.

More broadly, performing \acrshort{pca} involves the steps:
\begin{enumerate}
    \item Center the columns of $\mathbf{X}$ to have zero mean. 
    \item Find the singular values and corresponding singular vectors. The singular values and right singular vectors of $\mathbf{X}$ correspond to the eigenvalues and eigenvectors of the covariance matrix $\mathbf{S} = \frac{1}{m-1}\mathbf{X}^{\top}\mathbf{X}$.
    \item Project $\mathbf{X}$ onto the hyperplane spanned by the $\ell$ largest eigenvectors of the covariance matrix $\mathbf{S}$. The symmetry of the covariance matrix $\mathbf{S}$ implies that all of its eigenvectors are orthogonal. 
\end{enumerate}

The complete projection step of \acrshort{pca} can be written as
\begin{equation}
    \mathbf{T} = \mathbf{XW}
    \label{eq:pca}
\end{equation}
 where $\mathbf{T} \in \mathbb{R}^{m \times n}$ is the \acrshort{pca} score matrix, $\mathbf{X} \in \mathbb{R}^{m \times n}$ is the data matrix, and $\mathbf{W} \in \mathbb{R}^{n \times n}$ is a coefficient matrix. The columns of $\mathbf{W}$ are called loadings and correspond to the eigenvectors of $\mathbf{X}^{\top}\mathbf{X}$ in descending order. 
 
Equation \ref{eq:pca} describes a linear transformation. The loadings form an orthogonal basis, which results in the columns of $\mathbf{T}$ being decorrelated \citep{bengio2014representation}. 
The first $\ell < n$ components of $\mathbf{T}$ form the $\ell$-dimensional representation of $\mathbf{X}$ preserving most variance and are denoted by $\mathbf{T}_\ell$. 

Applied data analytics problems nearly always have $\ell < m$, in which case \acrshort{ols} can be used for regressing the lower dimensional representation $\mathbf{T}$ versus $\mathbf{y}$ in an additional step. The combination of these two methods is known as \acrfull{pcr}. Combining  \eqref{eq:OLSclosedform} and \eqref{eq:pca} leads to
\begin{equation}
    \boldsymbol{\beta}_\ell = (\mathbf{T_{\ell}}^{\top}\mathbf{T}_{\ell})^{-1}\mathbf{T_{\ell}}^{\top}\mathbf{y}
    \label{eq:pcr_lowdim}
\end{equation}
where $\ell$ is the number of principal components. The regression coefficients $\boldsymbol{\gamma}_\ell$ correspond to the lower dimensional space based on $\ell$ principal components. Consequently, the regression coefficients in the original space are constructed by 
\begin{equation}
    \boldsymbol{\beta} = \mathbf{W}_\ell \boldsymbol{\beta}_\ell
    \label{eq:pcr_}
\end{equation}
where $\mathbf{W}_\ell$ denotes the first $\ell$ columns of the matrix $\mathbf{W}$. 

\subsection{PLS}

\acrfull{pls} aims to find lower dimensional representations of $\mathbf{X}$ and $\mathbf{Y}$ and is not restricted to scalar objectives. While \acrshort{pca} performs dimensionality reduction in an unsupervised way, \acrshort{pls} incorporates information about the target $\mathbf{Y}$ in the dimensionality reduction scheme. In general, the governing equations for \acrshort{pls} are
\begin{align}
    \mathbf{X} &= \mathbf{TP}^{\top} + \mathbf{E} \\
    \mathbf{Y} &= \mathbf{UQ}^{\top} + \mathbf{F}    
    \label{eq:pls}
\end{align}
where the matrix $\mathbf{X} \in \mathbb{R}^{m\times n}$ is the data matrix, and $\mathbf{Y} \in \mathbb{R}^{m \times p}$ is the matrix of responses, $\mathbf{T} \in \mathbb{R}^{m \times l}$ is the score matrix, $\mathbf{P} \in \mathbb{R}^{n \times l}$ is the loading matrix, and $\mathbf{E} \in \mathbb{R}^{m\times n}$ is the residual matrix corresponding to $\mathbf{X}$. Similarly for $\mathbf{Y}$, the matrix $\mathbf{U} \in \mathbb{R}^{m \times l}$ is its score matrix, $\mathbf{Q} \in \mathbb{R}^{p \times l}$ is its loading matrix, and $\mathbf{F} \in \mathbb{R}^{m\times p}$ is its residual matrix.

In data analytics applications, the case $p = 1$ occurs frequently and is known as \acrshort{pls}1. The next paragraph introduces \acrshort{pls}1 with $\mathbf{y}$ being an single column vector. Readers interested in more details on other variants of \acrshort{pls} are directed to other references  \citep{chiang2000fault, WOLD2001109, PLSlatent, BOULESTEIX06plsgenom}. 

Similarly to \acrshort{pca}, \acrshort{pls} performs a linear transformation on the input $\mathbf{X}$, with
\begin{equation}
    \mathbf{T} = \mathbf{XW}.
\end{equation}
From an optimization perspective, \acrshort{pls} maximizes the sample covariance between the $\mathbf{X}$ scores and the responses \citep{WOLD2001109, BOULESTEIX06plsgenom}. The calculation of the first component can be written in the form of the unconstrained optimization
\begin{equation}
    \mathbf{w_1} = \argmax_\mathbf{W}
    \frac{\mathbf{w}^{\top}\mathbf{X}^{\top}\mathbf{yy}^{\top}\mathbf{Xw}}{\mathbf{w}^{\top}\mathbf{w}}.
    \label{eq:PLS_opt}
\end{equation}
The regression part of \acrshort{pls} is identical to \eqref{eq:pcr_lowdim}, but using $\mathbf{T_{\ell}}$ estimated via the \acrshort{pls} algorithm.

%% file: 3DemonstratorDesign.tex
\section{Demonstrator Design}
\label{sec:Demonstrator}

\begin{figure}[H]
    \centering
    \begin{tikzpicture}
        \node[above right, inner sep=0](image) at (0,0) {
            \includegraphics[trim=4mm 3mm 2.5mm 6mm, clip,     width=0.85\textwidth]{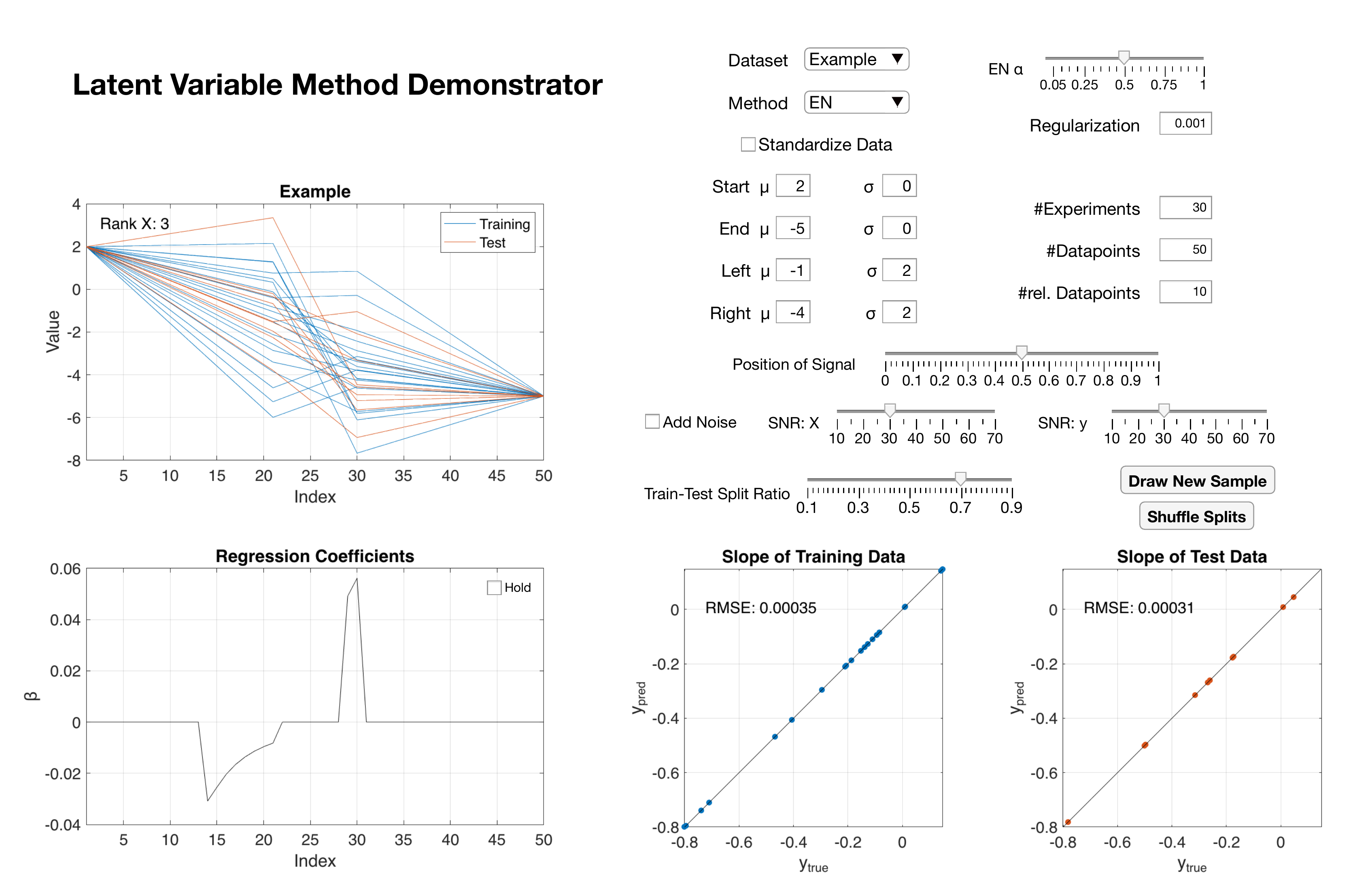}
        };
        
        \begin{scope}[
        x={($0.1*(image.south east)$)},
        y={($0.1*(image.north west)$)}]
        \ifnum \value{develop}>0
        	\draw[lightgray,step=0.5] (image.south west) grid (image.north east);
        
        	\foreach \x in {0,1,...,10} { \node [below] at (\x,0) {\x}; }
        	\foreach \y in {0,1,...,10} { \node [left] at (0,\y) {\y};}
        \fi
        \draw[thick,red] (0,0.05) rectangle (4.1,4.1) 
            (0,4.1)node[below right,black,fill=red]{\scriptsize 5};
        \draw[thick,red] (0,4.25) rectangle (4.1,8.3)
            (0,8.3)node[below right,black,fill=red]{\scriptsize 4};
        \draw[thick,red] (4.5,0.05) rectangle (9.8,4.0)
            (4.5,4.0)node[below right,black,fill=red]{\scriptsize 6};
        \draw[thick,red] (4.5, 8.4) rectangle (6.9,9.8)
            (4.5,9.8)node[below right,black,fill=red]{\scriptsize 1};
        \draw[thick,red] (7.1,8.4) rectangle (9.8,9.8)
            node[below left,black,fill=red]{\scriptsize 2};
        \draw[thick, red] (4.5,4.1) -- (7.8,4.1) -- (7.8, 4.95) -- (9.8, 4.95)-- (9.8, 8.3)-- (4.5, 8.3) -- cycle
            (4.5,8.3)node[below right,black,fill=red]{\scriptsize 3};
        \end{scope}
    \end{tikzpicture}
    \caption{Layout of \acrfull{lavade}}
    \label{fig:lavade_layout}
\end{figure}

The \acrfull{lavade} software is designed to evaluate and compare the regression techniques described in Section \ref{sec:RegressionTechniques} on datasets of different complexity. The \acrfull{gui} of the software is a single window, with drop-down menus, sliders, buttons, and axis objects (Fig.\ \ref{fig:lavade_layout}). Depending on the selection of the dataset and the regression technique, the relevant elements on the window become visible or invisible, reducing user distraction. 
\newcounter{develop}
\setcounter{develop}{0}

The six elements in the GUI are
\begin{enumerate}
    \item Dataset and Method selection drop-down menu, and check box, to select whether inputs shall be standardized\footnote{Standardization refers to centering the columns and scaling to have unit variance.}
    \item Hyperparameters for the respective regression method 
    \item Modification of data: Data generation parameters (``Example case''), addition of white Gaussian noise
    \item Visualization of the data, training, and test data, color-coded
    \item Visualization of regression coefficients     
    \item Regression results on the training and test data, respectively
\end{enumerate}

On the backend, the software consists of two MATLAB classes. The ``lavade'' class inherits from the ``AppBase'' class, a generic MATLAB app class and contains all the elements of the \acrshort{gui} and general variables associated with the functionality of the software. The ``data'' class is a superclass from which all the individual data set classes inherit. The constructor of the individual dataset classes loads or generates the data. Furthermore, the data class contains methods to perturb the signal by noise, and to split the data. This structure helps to implement new datasets quickly.

%% file: 4Datasets.tex
\section{Datasets}
\label{sec:datasets}

The \acrshort{lavade} software includes an illustrative numerical example data generator and three real-world datasets. The datasets can also be downloaded separately from the corresponding GitHub repository.

\begin{figure}
    \centering
    \begin{tikzpicture}
        \node[above right, inner sep=0](image) at (0,0) {
            \includegraphics[scale=0.7]{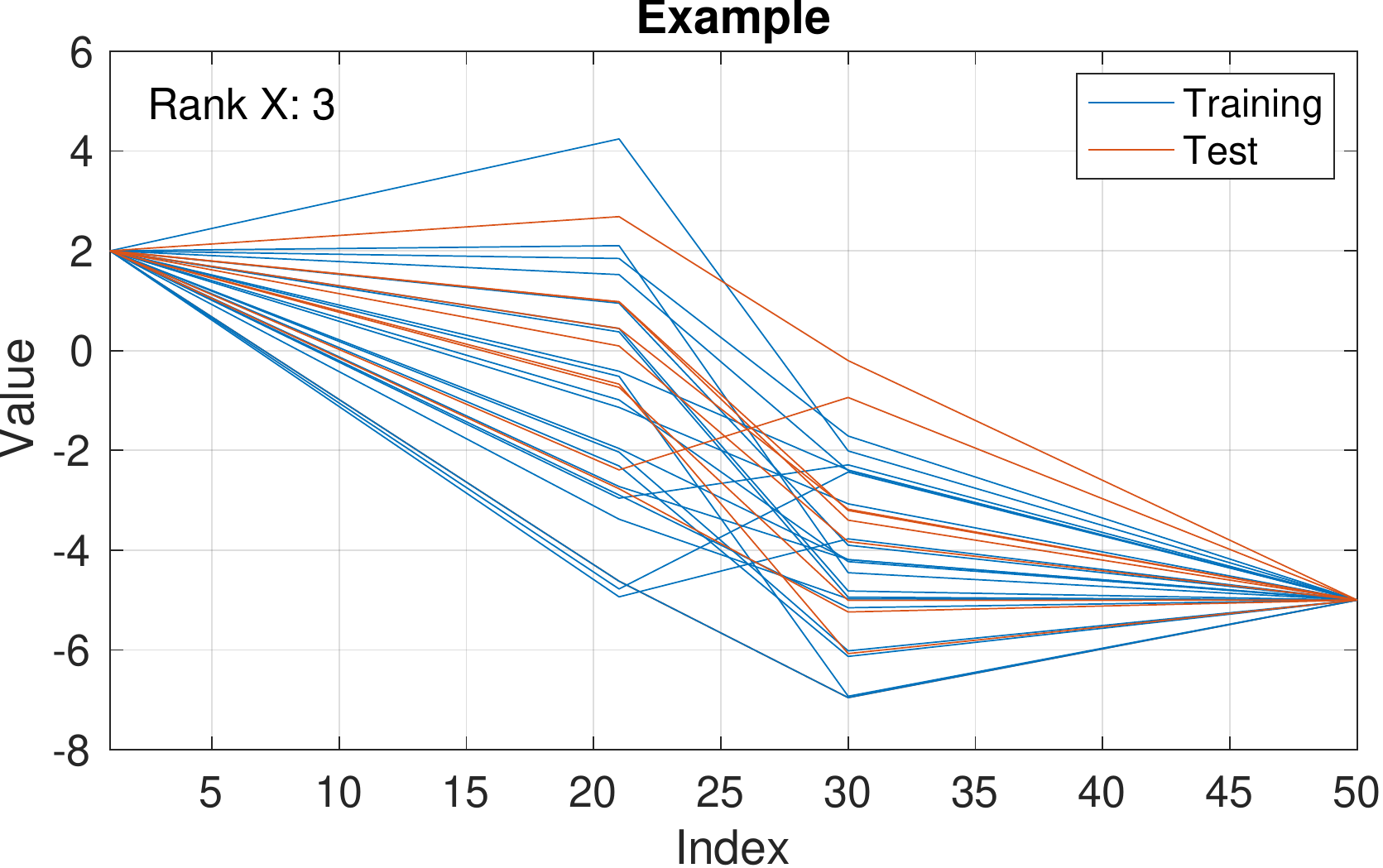}
        };
        
        \begin{scope}[
        x={($0.1*(image.south east)$)},
        y={($0.1*(image.north west)$)}]
        \ifnum \value{develop}>0
        	\draw[lightgray,step=0.5] (image.south west) grid (image.north east);
        
        	\foreach \x in {0,1,...,10} { \node [below] at (\x,0) {\x}; }
        	\foreach \y in {0,1,...,10} { \node [left] at (0,\y) {\y};}
        \fi
        \draw[thick,red] (0.828,1.4) rectangle (4.46,9.38) 
            (0.828,1.4)node[above right,black,fill=red]{\scriptsize Start}
            (4.46,1.4)node[above left,black,fill=red]{\scriptsize Left};
        \draw[thick,red] (0.828+5.33,1.4) rectangle (4.46+5.33,9.38) 
            (0.828+5.33,1.4)node[above right,black,fill=red]{\scriptsize Right}
            (4.46+5.33,1.4)node[above left,black,fill=red]{\scriptsize End};
        \draw[thick,black!60!green] (4.5,9.38) rectangle (0.788+5.33,1.4)
            (5.2,1.9)node[black!60!green, text width=1.5cm]{\scriptsize \begin{tabular}{c} \textbf{Relevant} \\ \textbf{Section} \end{tabular}};
        \end{scope}
    \end{tikzpicture}
    \caption{Default case: $\sigma_{\textrm{start}}=\sigma_{\textrm{end}}=0$, $\sigma_{\textrm{left}}=\sigma_{\textrm{right}}=2$. ``Relevant section'' refers to the middle section among the three sections.} 
    \label{fig:DefaultData}
\end{figure}
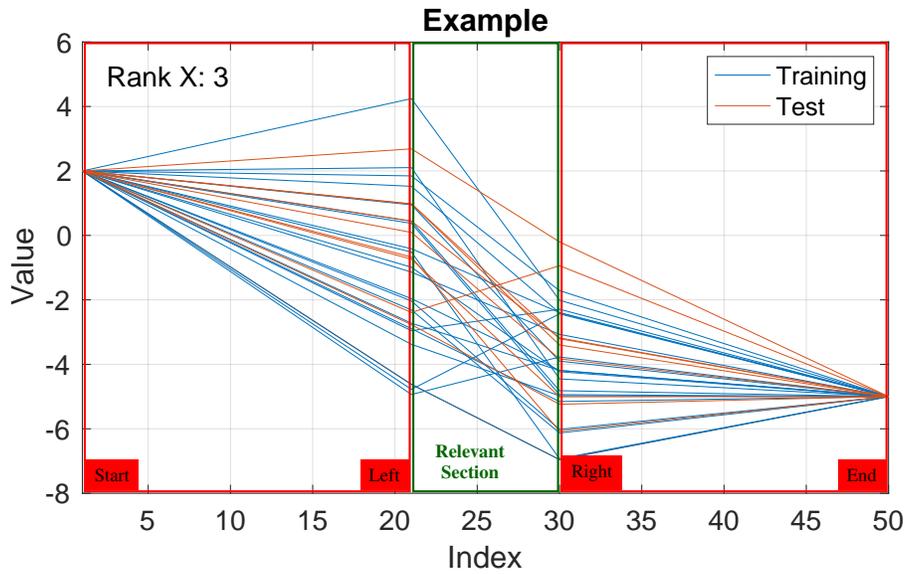

\paragraph{Example}
The motivation of the example dataset is to generate a dataset similar in character to process measurements, in which only a subset of the measurements have predictive power. The dataset is constructed by starting by defining four data points: start, final, and two intermediate data points (Fig.\ \ref{fig:DefaultData}). The four data points are drawn from a Gaussian distribution with defined means and standard deviations. The output responses $\mathbf{y}$ are the (average) slopes in the relevant section. 

\setcounter{develop}{0}

Figure \ref{fig:DefaultData} shows the default case in which the parameters of the four Gaussian distributions are $\mu_{\textrm{start}}=2$, $\mu_{\textrm{left}}=-1$, $\mu_{\textrm{right}}=-4$,
$\mu_{\textrm{end}}=-5$, and $\sigma_{\textrm{start}}=\sigma_{\textrm{end}}=0$, $\sigma_{\textrm{left}}=\sigma_{\textrm{right}}=2$.  The data are within three sections. The slope in the relevant section can also be inferred from data outside of the relevant section due to $\sigma_{\textrm{start}}=\sigma_{\textrm{end}}=0$.

\begin{figure}[H]
    \centering
    \includegraphics[scale=0.7]{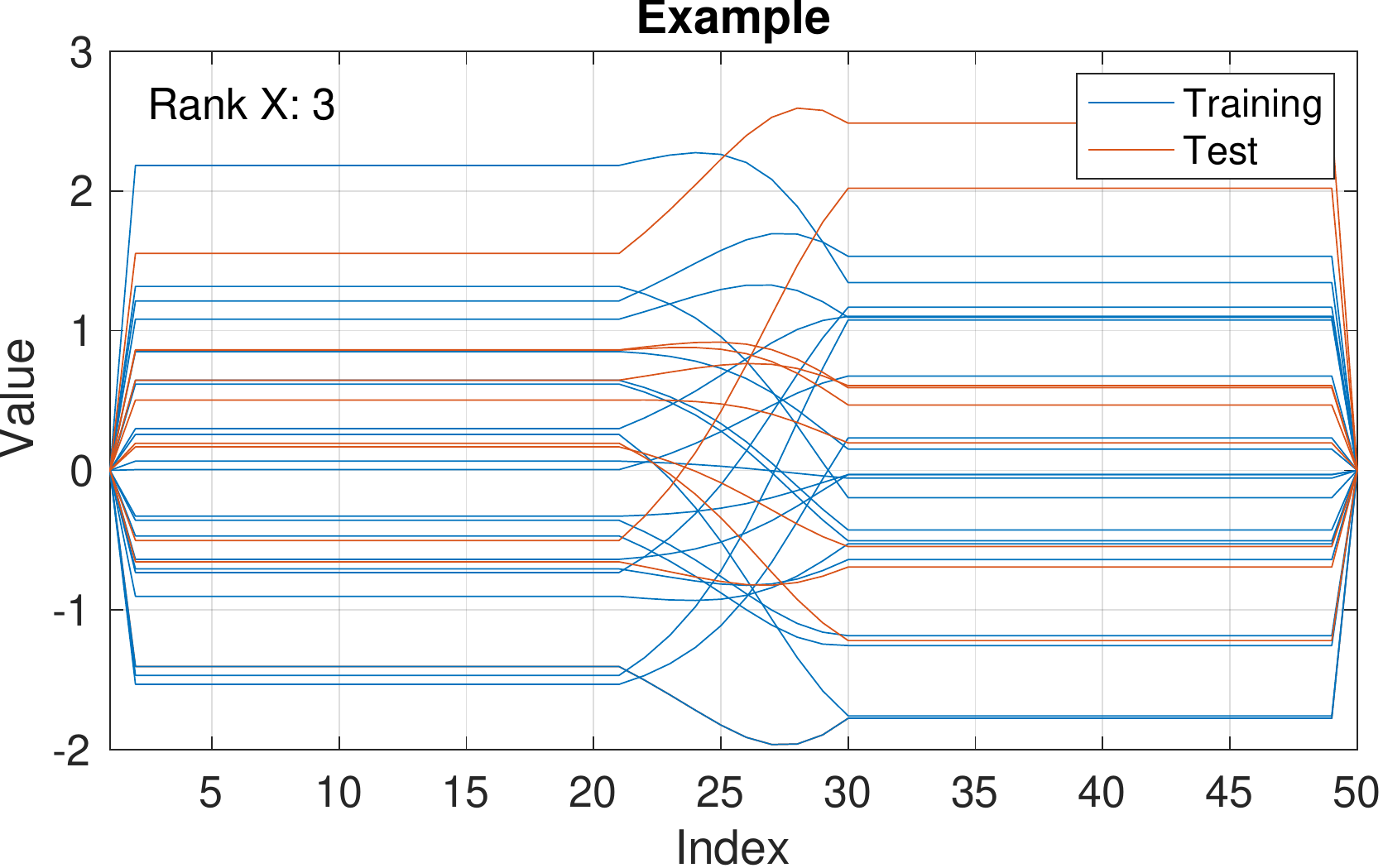}
    \caption{Default standardized case: $\sigma_{\textrm{start}}=\sigma_{\textrm{end}}=0$, $\sigma_{\textrm{left}}=\sigma_{\textrm{right}}=2$, standardized columns.} 
    \label{fig:DefaultStandData}
\end{figure}

This dataset illustrates the common situation in which the data are highly correlated while not having any datapoints repeated, while being simple enough to easily understand the dataset and the model being constructed, and be able to interpret the results.

Figure \ref{fig:DefaultStandData} shows the result of standardizing the data in Fig.\ \ref{fig:DefaultData}. For each data vector, all datapoints left of the relevant section have the same value and carry the same information, except for the first and the last datapoints. The same holds true for the datapoints right of the relevant section. 

\begin{figure}[H]
    \centering
    \includegraphics[scale=0.7]{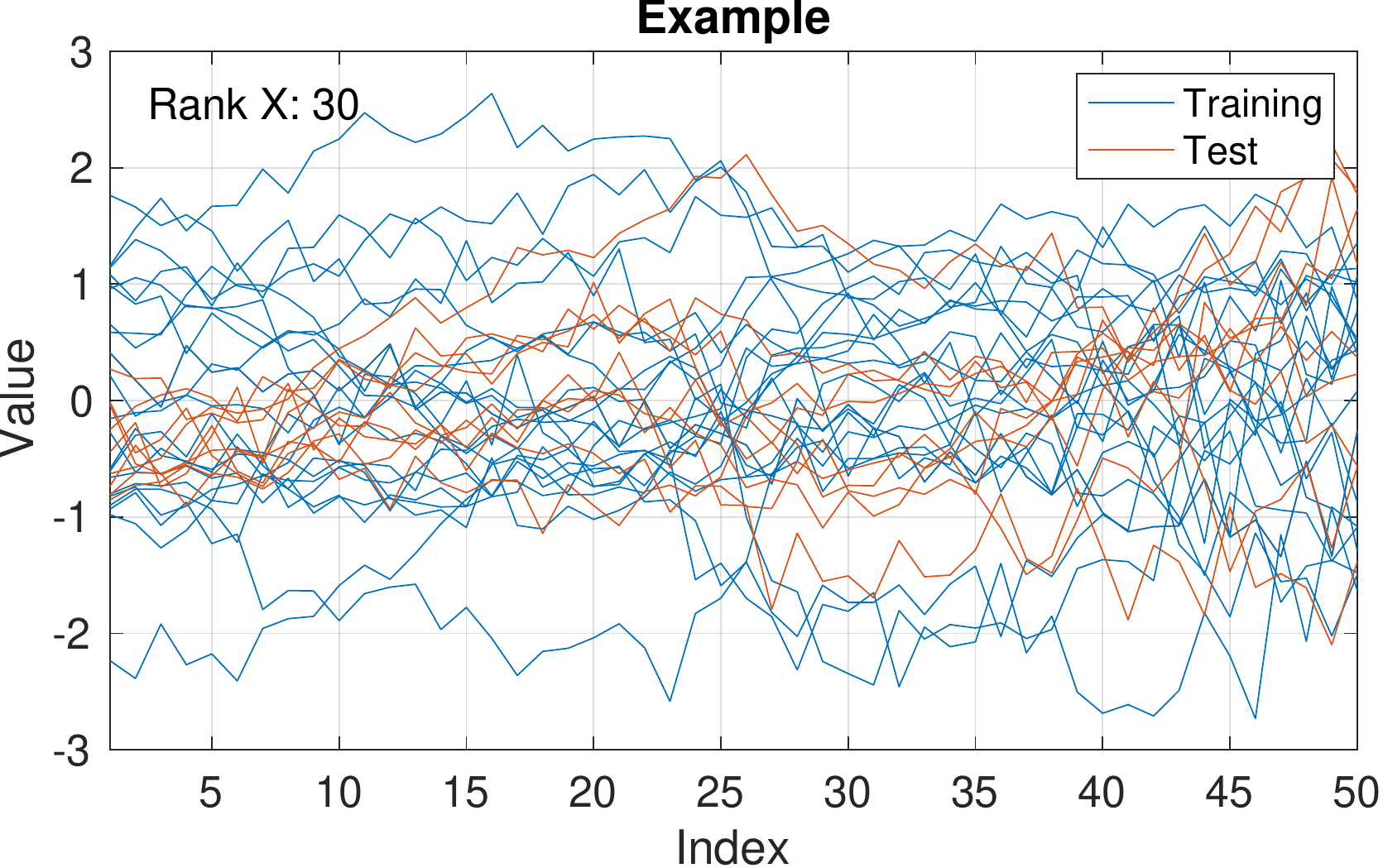}
    \caption{Noisy case: $\sigma_{\textrm{start}}=\sigma_{\textrm{end}}=2$, $\sigma_{\textrm{left}}=\sigma_{\textrm{right}}=2$, $\text{\acrshort{snr}}=20$ for all sections.} 
    \label{fig:NoisyData}
\end{figure}

A noisy dataset adds white Gaussian noise with defined  \acrfull{snr} to the matrix $\mathbf{X}$ and vector of responses $\mathbf{y}$ (Fig.\ \ref{fig:NoisyData}). By varying the \acrfull{snr}, the user can assess how noise affects the model training and model predictions. 
This example dataset encourages the user to develop a general understanding of model types, hyperparameters, noise, training, and prediction accuracy on a simple dataset inspired by common process data analytics problems. 
 
\paragraph{ATR-FTIR Spectra}
This dataset consists of Attenuated Total Reflectance-Fourier Transform Infrared (ATR-FTIR) spectra that were recorded and analyzed in \cite{fujiwara2002paracetamol}. The model objective is to predict the paracetamol concentration $C$ associated with each spectrum. The dataset consists of spectra collected for six values of the concentration (six ``groups''), with varying temperature. For simplicity, we omit the temperature readings and use only the ATR-FTIR spectra as model input.

\begin{figure}[H]
    \centering
    \includegraphics[scale=0.7]{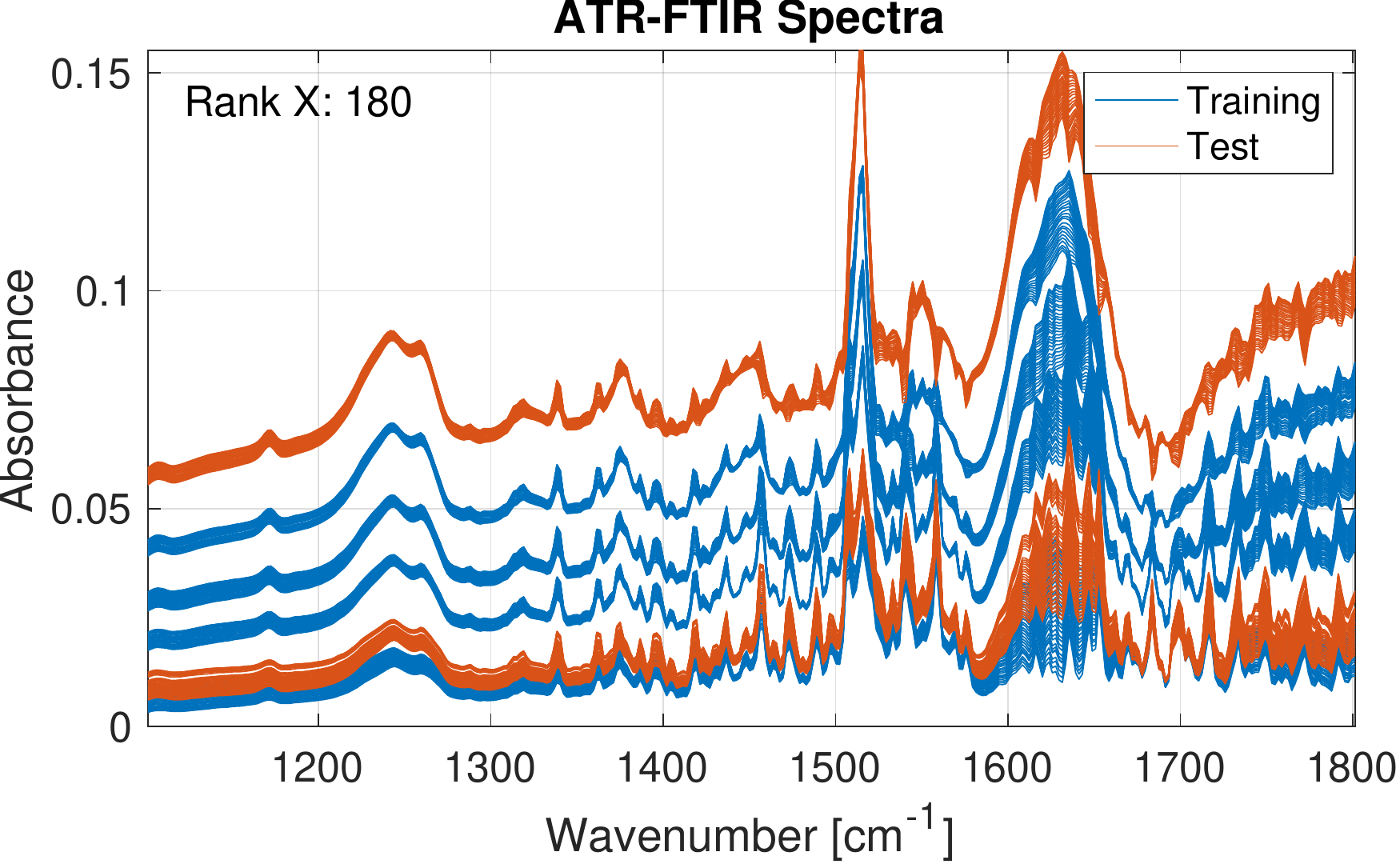}
    \caption{Visualization of ATR-FTIR spectra for paracetamol in aqueous solution.} 
    \label{fig:visparacetamol}
\end{figure}

Each spectrum is highly correlated with the other spectra, and the absorbances within each spectrum are highly correlated. Each group of spectra is evenly spaced from adjacent groups of spectra except for the group of spectra at $C=0.014\, \text{g/g}$, which is the second group from the bottom in Fig.\ \ref{fig:visparacetamol}. Because the rest of the data appears to have roughly a linear relationship, it is expected that much of the $C=0.014\, \text{g/g}$ data are biased, most likely by some experimental error. As such, the data at $C=0.014\, \text{g/g}$ are placed in the test dataset as the default. This data assignment leads to more conservative test errors and prevents the model from learning an aphysical correlation due to this bias. 
The strategy of splitting of this dataset can be chosen in the \acrshort{gui} to be either fully at random or such that groups of spectra roughly corresponding to the same concentration are always part of only the training or the test dataset. This latter respects the fact that, for spectral data, the variation of data between groups is typically larger than the variation of data within a group \citep{sun2021smart}.
Given that error structure, placing the data from all groups into the training dataset, for example, would result in overfitting and overly confident/optimistic prediction errors for the test data \citep{Escobar2014Grouping}. 
Usually, data are standardized before applying ridge regression, lasso, and elastic net, to account for widely different ranges of values for each data type (e.g., a temperature could be more than 100 degrees for a system in which absorbance spans 0.01 absorbance units). However, this particular dataset consists only of spectra, and all readings have the same order of magnitude. In this case, skipping standardization can result in better models. Typically many absorbances are not associated with the molecule of interest and contain nearly all noise; standardization normalizes those absorbances to have the same magnitude of importance in the matrix $\mathbf{X}$ in the optimization objective. Thereby standardization feeds noise into the parameter estimation and degrades the quality of the model. Also, the potential benefits of standardization are lower for spectral data, as most absorbance values are already of a similar order of magnitude. 

\paragraph{Raman Spectra}
Raman scattering refers to the inelastic scattering of photons by matter. Raman spectra have high molecular specificity and are used to build models to predict molecular concentrations in solutions \citep{raman_recent_advances, raman_glucose}. A promising use case for this type of model is noninvasive glucose monitoring \citep{Raman_bioreactor_generic_raman}, which uses Raman spectra from different bioreactor runs to learn mappings to glucose concentrations from Nova FlexII cell culture measurements. This task is challenging due to noise and background signals in the Raman spectra and noise in the Nova FlexII readings. Furthermore, the biochemical reactions taking place in the bioreactor lead to correlations in the concentration changes of various metabolites. 
We removed outliers from the cell culture measurements of \cite{Dragana} by applying a moving median filter having a window size of three days and removing all values that were above three standard deviations. The Raman spectra are available in the software (1) without background removal, (2) background removal by the method of  \cite{LieberFlourFilter} using the Biodata MATLAB Toolbox implementation \citep{BiodataToolboxMatlab}, and (3) background removal with the MATLAB algorithm \texttt{msbackadj}. The range of Raman shifts was restricted to the range 400--1800 cm$^{-1}$ as suggested by \cite{insituRaman} and \cite{atlineRaman}.
\begin{figure}[H]
    \centering
    \includegraphics[scale=0.7]{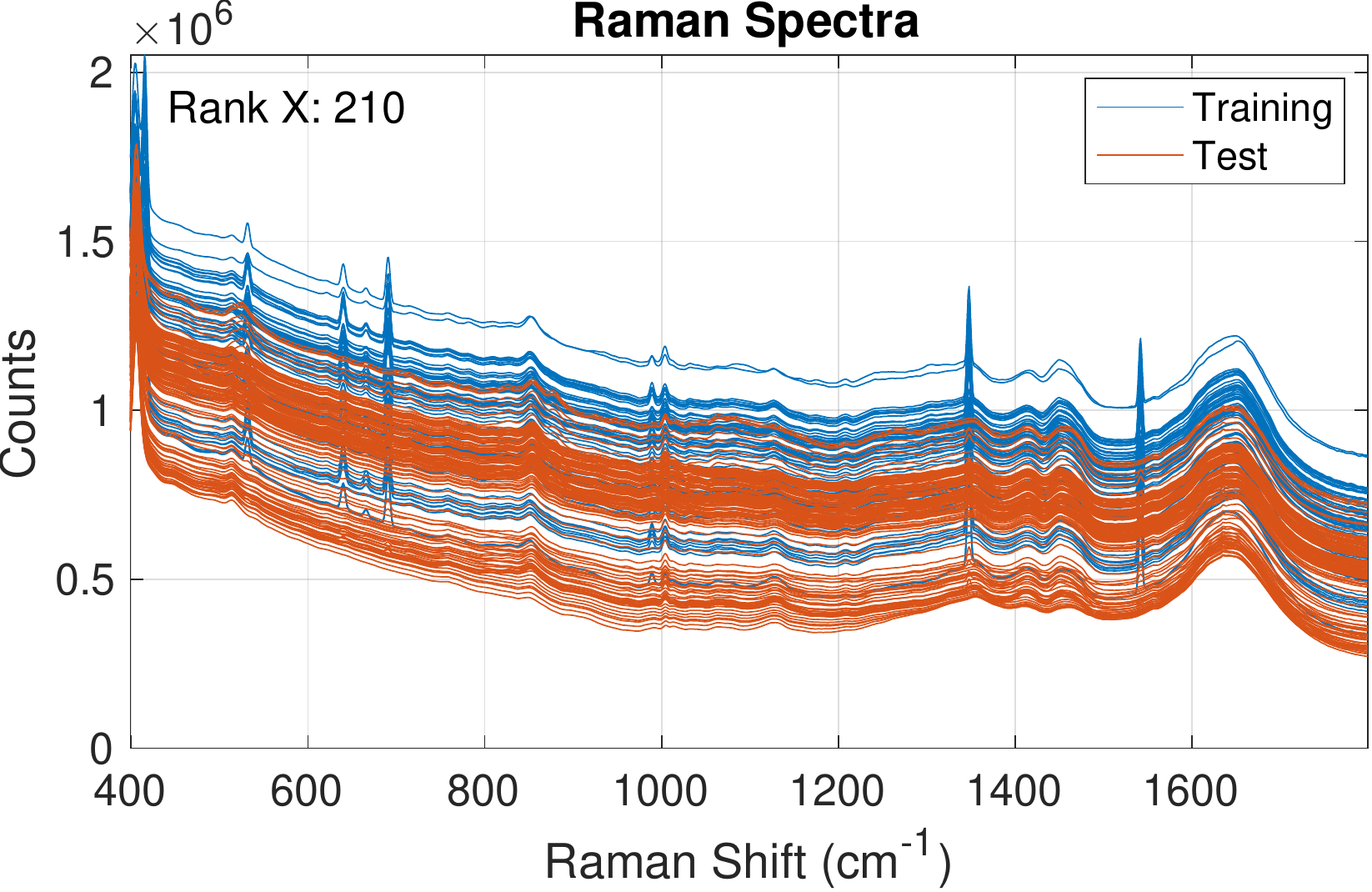}
    \caption{Visualization of Raman spectra without background removal from a bioreactor.} 
    \label{fig:raman_spectra}
\end{figure}

Figure \ref{fig:raman_spectra} shows 327 Raman spectra, each corresponding to a cell culture measurement. The cell culture measurements were collected at different times than the Raman spectra. The relatively slow concentration changes enabled the concentrations corresponding to the Raman spectra to be linearly interpolated from the cell culture measurements.

\paragraph{Lithium-Iron-Phosphate Batteries}
The Lithium-Iron-Phosphate (LFP) battery dataset contains cycling data for 124 batteries that have underwent widely varying charging protocols and discharged at a uniform current \citep{severson2019data}. The model objective is the cycle life, i.e., the number of cycles until the battery's capacity drops below $80\%$ of its nominal capacity.

\cite{severson2019data} showed that features based on the difference between the discharge capacity of voltage curves for two cycles, subsequently called $\Delta Q_{a-b}$, linearly correlate well with the logarithm of the cycle life. The cycle pair $a=100$ and $b=10$ was used in \cite{severson2019data}. While other cycle combinations also work well, the implemented input data $\mathbf{X}$ is $\Delta Q_{100-10}$ for this demonstrator. More information about the dataset and reasoning about the modeling objective can be found in \cite{severson2019data}. 

\begin{figure}[H]
    \centering
    \includegraphics[scale=0.7]{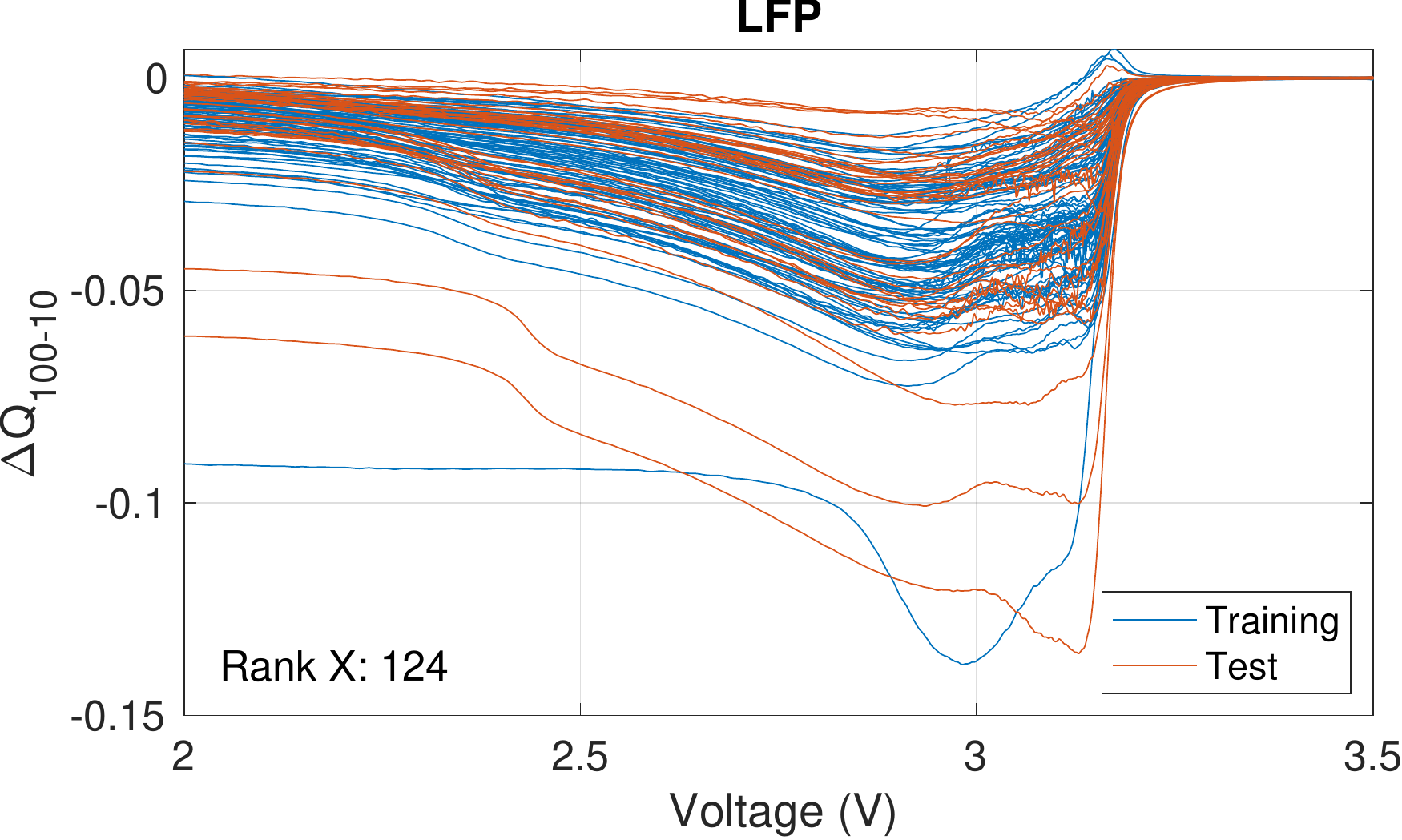}
    \caption{Visualization of the LFP dataset in the LAVADE software.} 
    \label{fig:vislfp}
\end{figure}

Some charging protocols were applied to only on a single cell, whereas others were used on up to nine different cells, leading to a grouping structure. In the \acrshort{lavade} software, the train-test split can be chose to be fully at random or randomly assigning groups such that cells with the same charging protocol are only part of either the training or the test dataset. The cycle life distribution of the dataset is non-Gaussian, with a heavy concentration of cells having a relatively low cycle life and only a few having longer cycle life. Depending on the split, models will tend to struggle to extrapolate to the longer-lived cells, leading to high variability in prediction errors calculated for the test set.

Furthermore, the batteries were cycled in three batches with different starting dates. Thus the cells were stored over different durations between manufacturing and cycling. The first batch was cycled approximately a year earlier than the last batch leading to prolonged calendar aging for some cells. This additional information is crucial when building robust data analytics models but is not elaborated on further in this context.

%% file: 5IllExamples.tex
\section{Illustrative Examples}
\label{sec:example}

This section illustrates the \acrshort{lavade} software by applying the various regression methods to the datasets. While small differences can arise when reproducing the results, due to the randomized splitting of the datasets, the observations made in this section are general. The regularization parameters in the software must be chosen by hand. For the following examples, we chose the hyperparameters to showcase interesting regression coefficients and prediction results.

\subsection{Example Dataset}
\noindent 
\begin{figure}[H]
\begin{center}
    \begin{minipage}{0.48\textwidth}
        \vspace{0.7cm}
        \includegraphics[width=\textwidth]{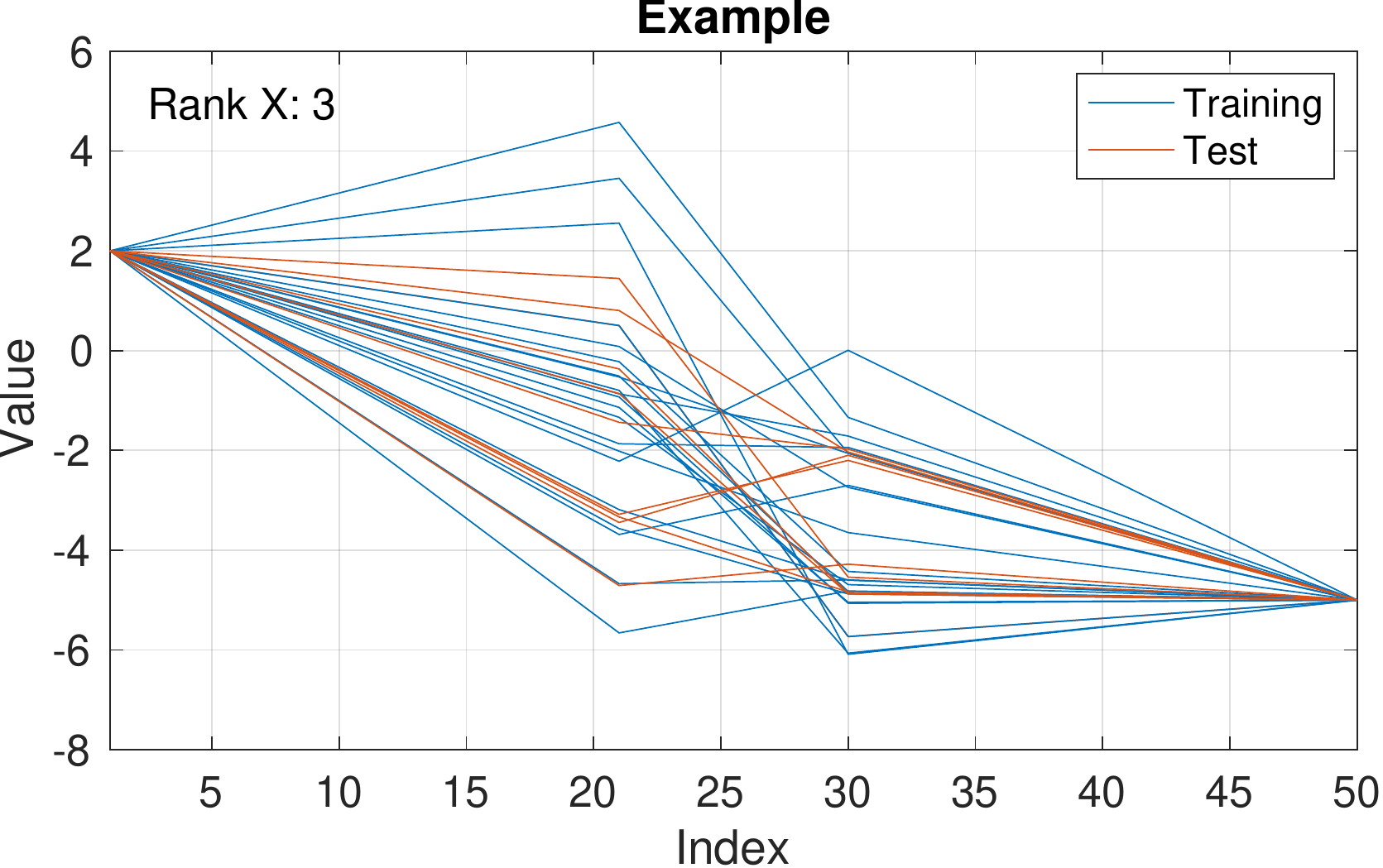}
        \label{fig:nature}
    \end{minipage}
    \vspace{0.02\textwidth}\\
    \begin{minipage}{\textwidth}
        \includegraphics[width=0.48\textwidth]{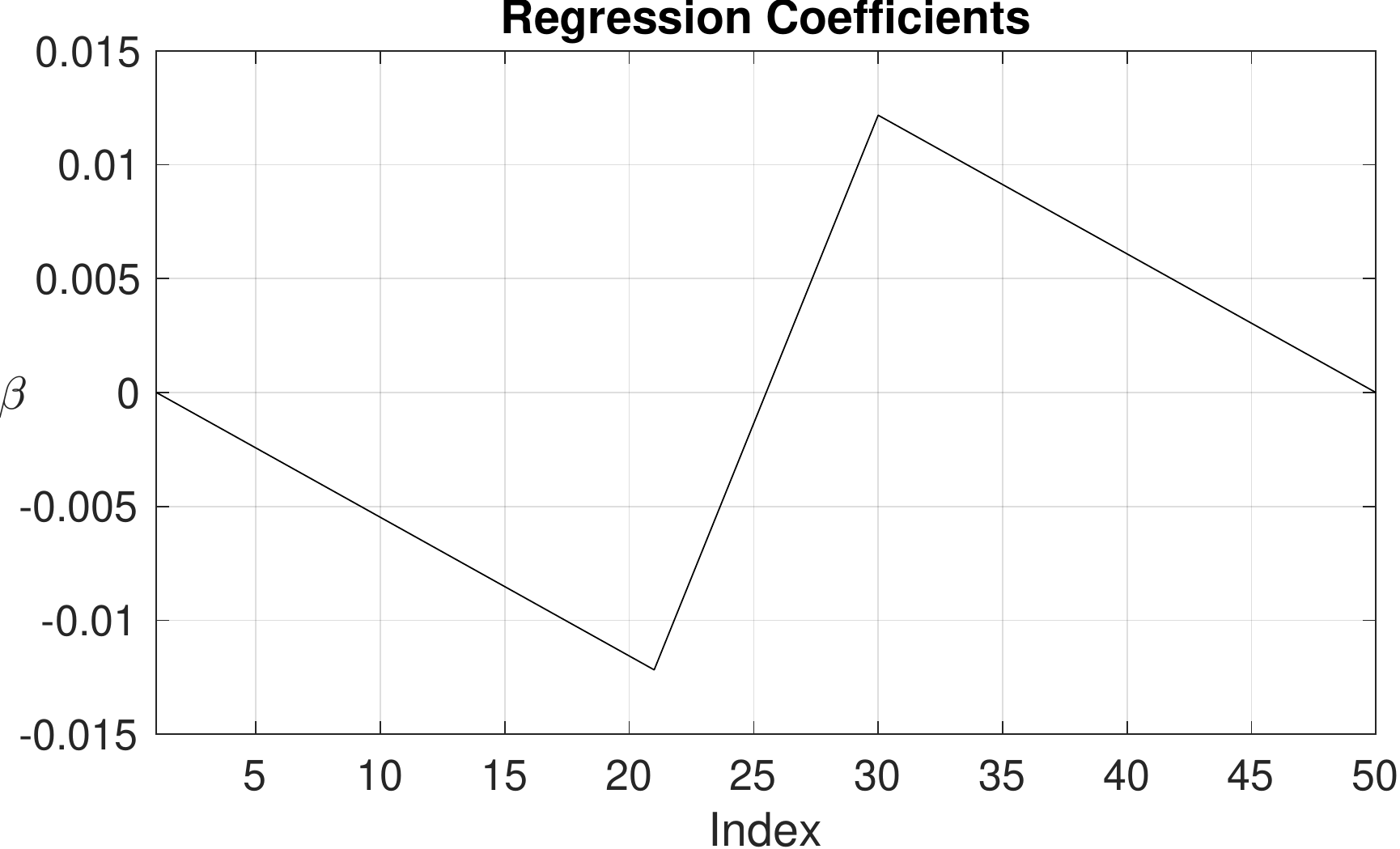}
        \hspace{0.02\textwidth}
        \includegraphics[width=0.48\textwidth]{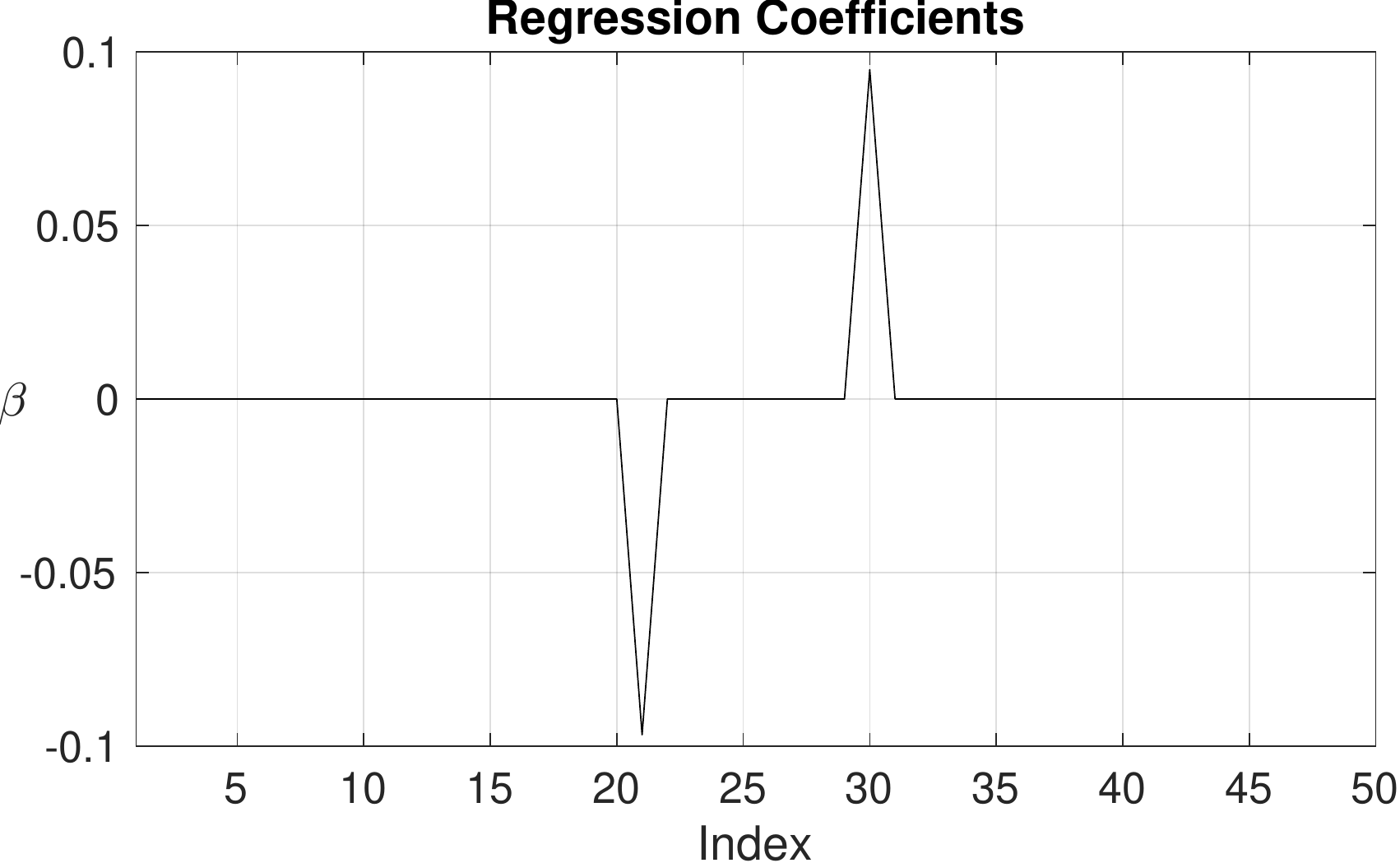}
    \end{minipage}
    \hspace{0.15cm}
\end{center}
\begin{picture}(0,0)
            \put(180,163){\footnotesize{(a) Noise-free example dataset}}
            \put(50,3){\footnotesize{(b) PLS and RR regression coefficients}}
            \put(310,3){\footnotesize{(c) lasso regression coefficients}}
\end{picture}
    \caption{Noise-free example dataset and regression coefficients for PLS, RR, and lasso, with hyperparameters and RMSEs reported in Table \ref{tab:ilex_ex}.}
    \label{fig:ilex_ex}
\end{figure}
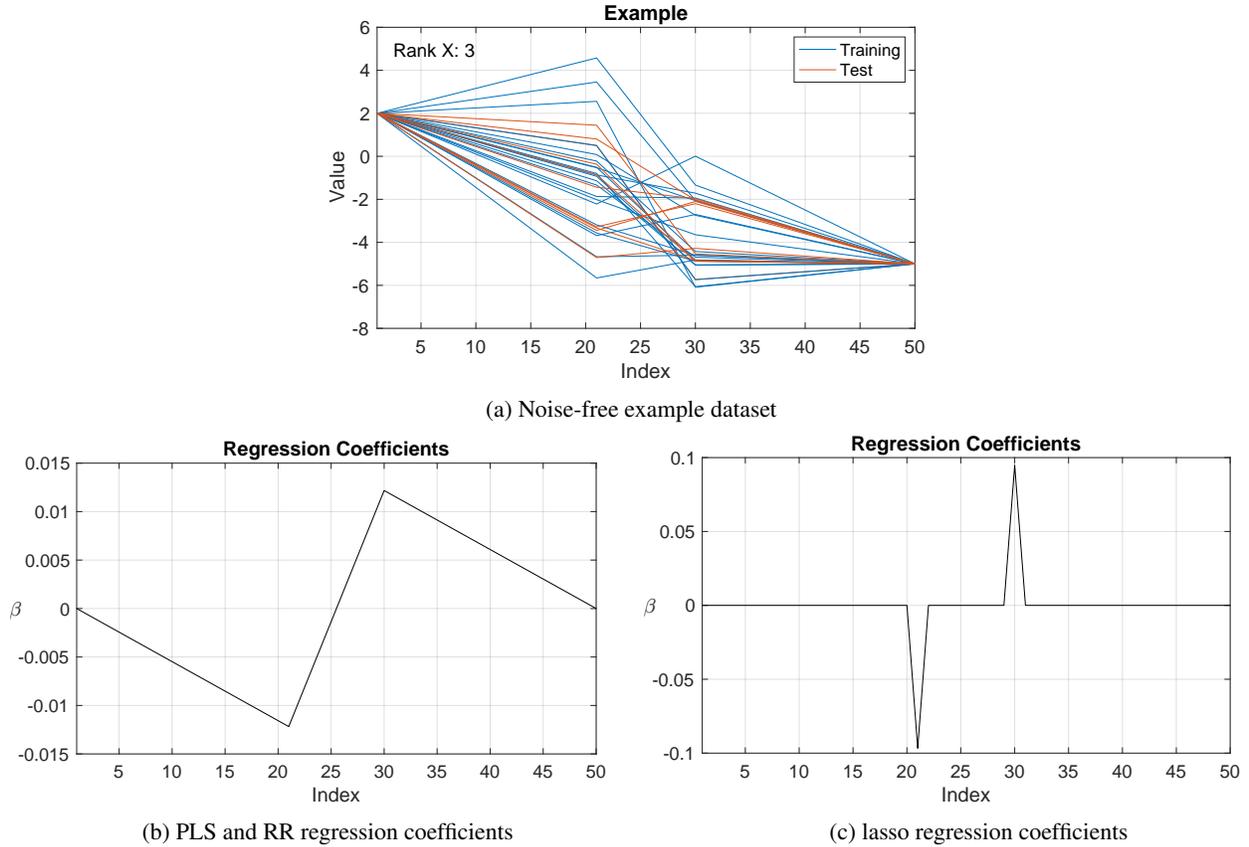
\begin{table}[H]
    \centering 
    \begin{tabular}{|c|c|c|c|} \hline
         Regression Method&\begin{tabular}{c}Regression\\ Coefficients\end{tabular} & RMSE Training & RMSE Test\\
        \hline \hline
         PLS  (2 components) & Fig.\ \ref{fig:ilex_ex}b & $1.4$$\times$$10^{-16}$ & $7.3$$\times$$10^{-17}$ \\
        \hline
        RR  ($\lambda=1$$\times$$10^{-6}$) &Fig.\ \ref{fig:ilex_ex}b &  $4.6$$\times$$10^{-10}$ & $4.6$$\times$$10^{-10}$ \\
        \hline
        lasso ($\lambda=0.015$) & Fig.\ \ref{fig:ilex_ex}c & 0.011 & 0.011 \\ \hline
    \end{tabular}
    \hspace{0.15cm}
    \caption{Regression methods, hyperparameters, regression coefficients, and RMSE for the training and test datasets for the noise-free example.}
    \label{tab:ilex_ex}
\end{table} 
The regression results for the noise-free example dataset illustrate several important properties of the implemented methods. \acrshort{pls} and \acrshort{rr} result in nearly identical regression coefficients in this ideal dataset (Fig.\ \ref{fig:ilex_ex}b). Due to the mathematical similarities between \acrshort{pcr}, \acrshort{pls}, and \acrshort{rr} \citep{DEJONG1994179, sundberg1993continuum_rr, stone1990CR_PLS_PCR} regression coefficients are expected to have certain similarities given appropriate parameters. \Acrshort{lasso} leads to sparse regression coefficients that are intuitive; an observant human would also recognize that all of the data points in Fig.\ \ref{fig:ilex_ex}a can be determined from the difference in values between the 30$^{\textrm{th}}$ and 21$^{\textrm{st}}$ measurements.\footnote{Or, equivalently, by the slope of the line in the middle region computed from the maximum and minimum values.} Table \ref{tab:ilex_ex} shows that the three methods investigated here lead to regression coefficients that result in accurate model predictions, with PLS and lasso being the most and least accurate, respectively, given the chosen regularization parameters. The regression coefficients, however, have very different shapes (cf.\ Fig.\ \ref{fig:ilex_ex}bc). Because the methods result in very similar predictions, the differences between the regression coefficients must be within or very close to the nullspace of the data, $\boldsymbol{N}(\mathbf{X})$.

\subsection{FTIR Spectral Data}

\noindent
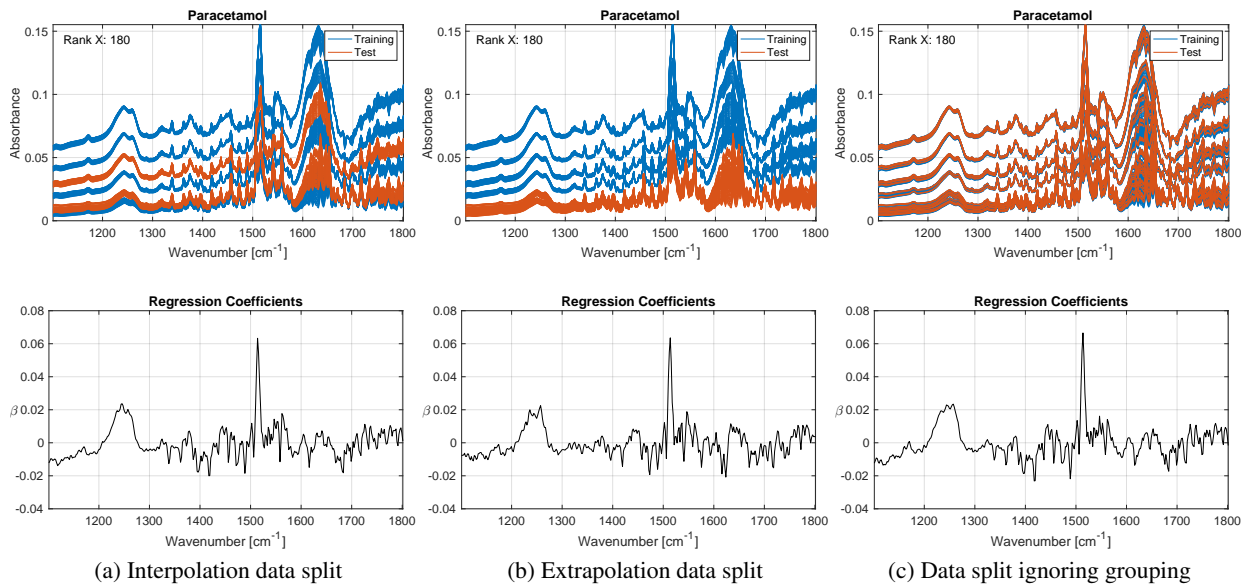
\begin{figure}[H]
\begin{minipage}{0.327\textwidth}
    \includegraphics[width=\textwidth]{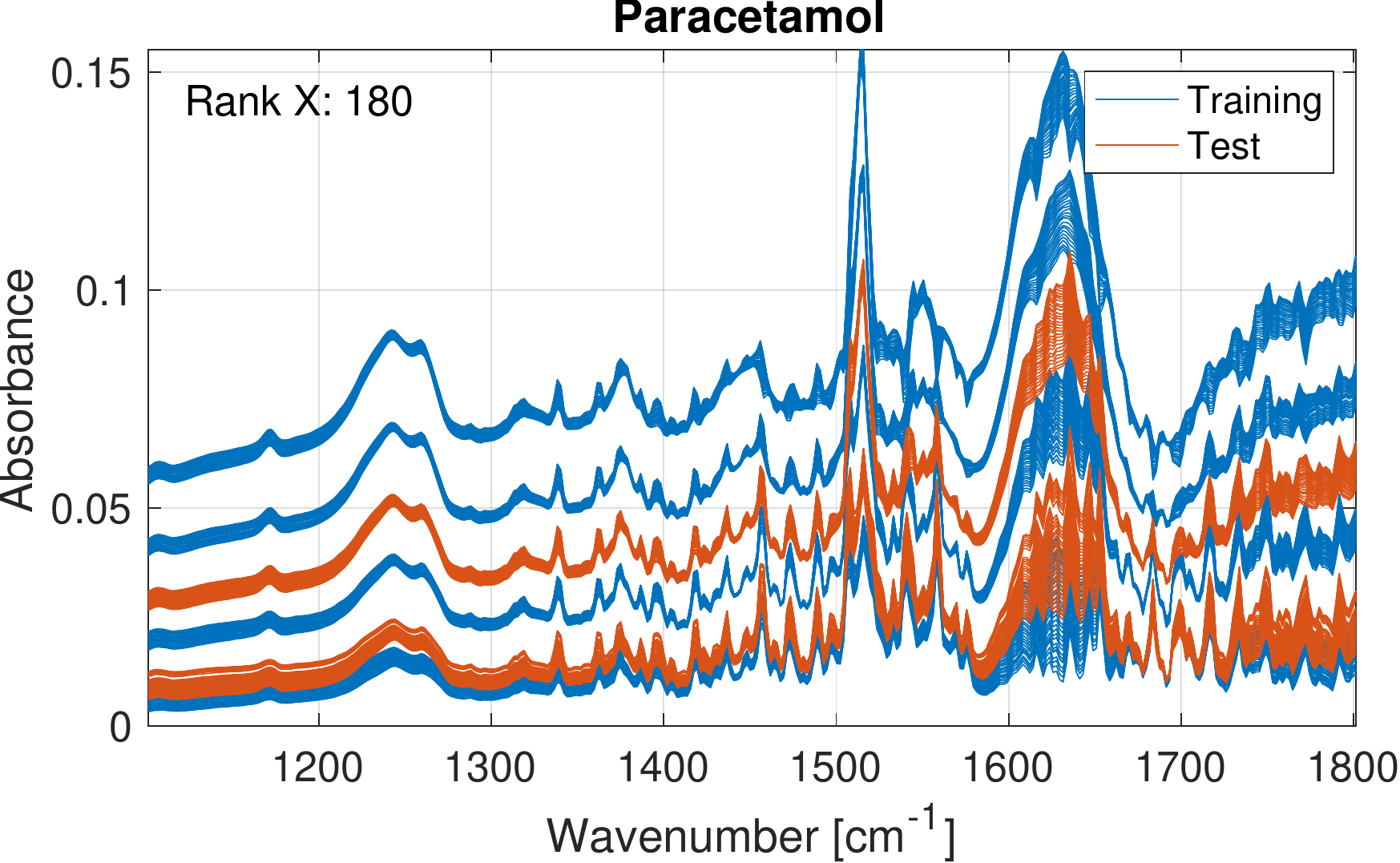}
    \vspace{0.01\textwidth}\\
    \includegraphics[width=\textwidth]{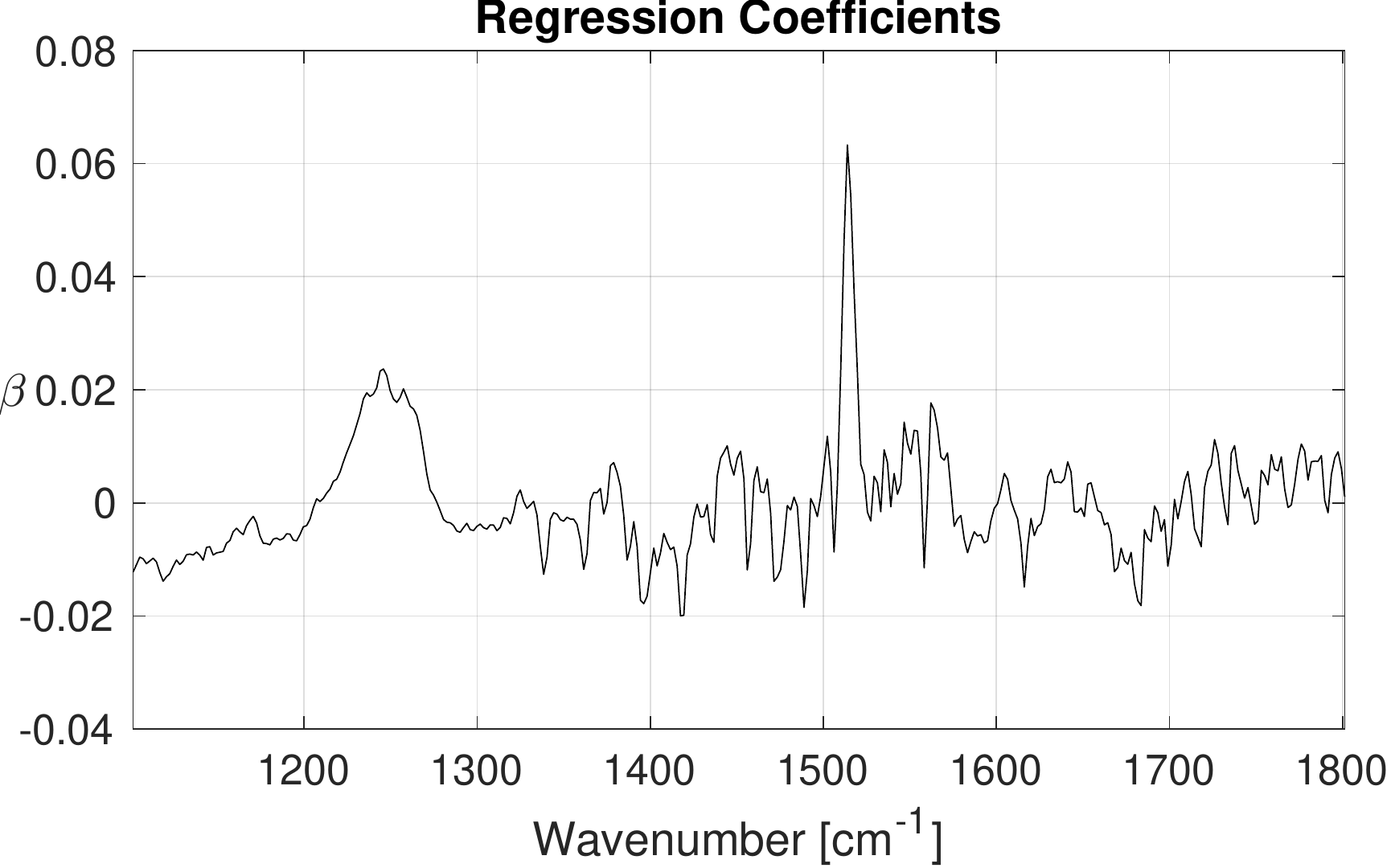}
    \begin{picture}(0,0)
        \put(32,0){\footnotesize{(a) Interpolation data split}}
    \end{picture}
\end{minipage}
\begin{minipage}{0.327\textwidth}
    \includegraphics[width=\textwidth]{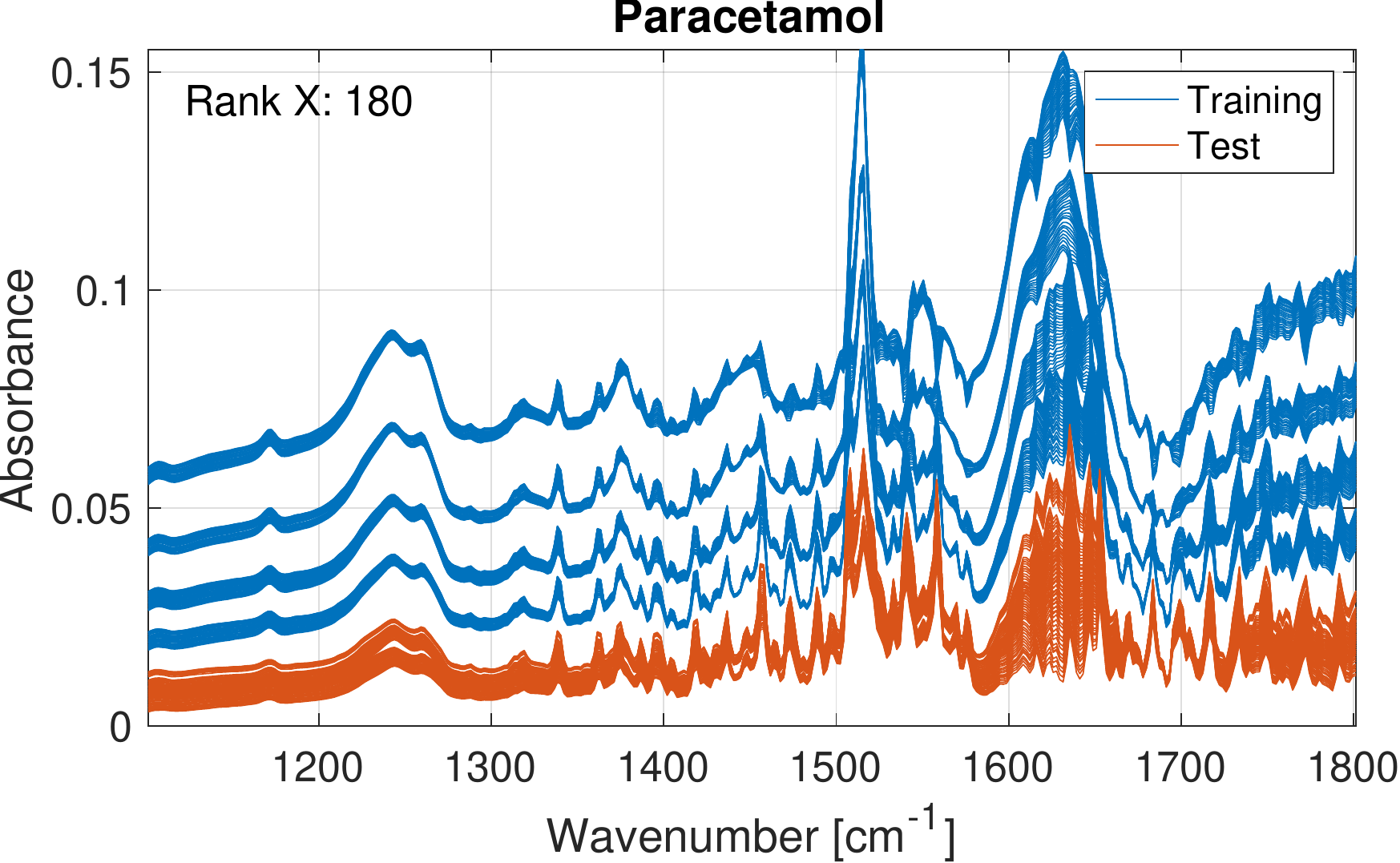}
    \vspace{0.01\textwidth}\\
    \includegraphics[width=\textwidth]{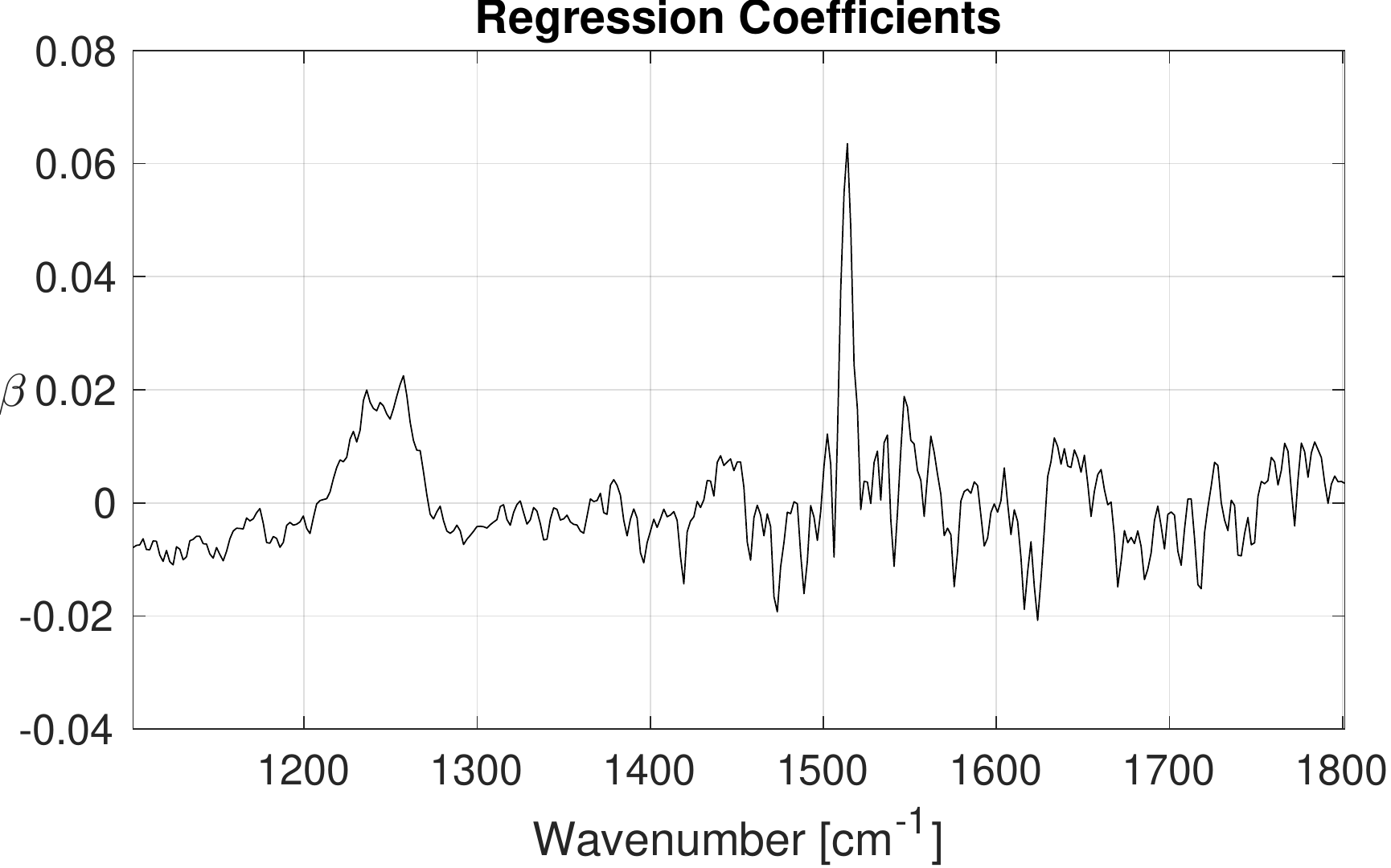}
    \begin{picture}(0,0)
        \put(32,0){\footnotesize{(b) Extrapolation data split}}
    \end{picture}
\end{minipage}
\begin{minipage}{0.327\textwidth}
    \includegraphics[width=\textwidth]{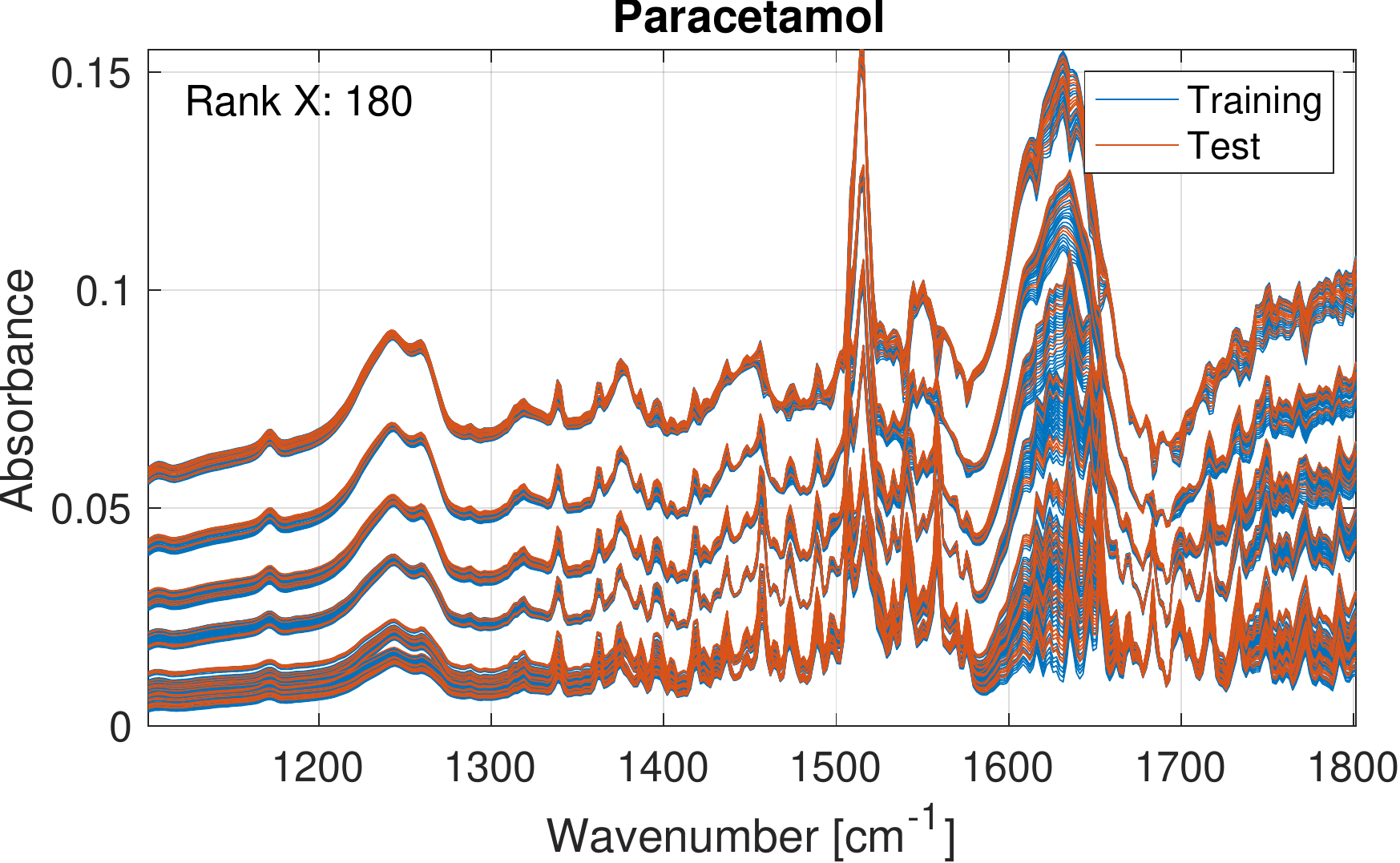}
    \vspace{0.01\textwidth}\\
    \includegraphics[width=\textwidth]{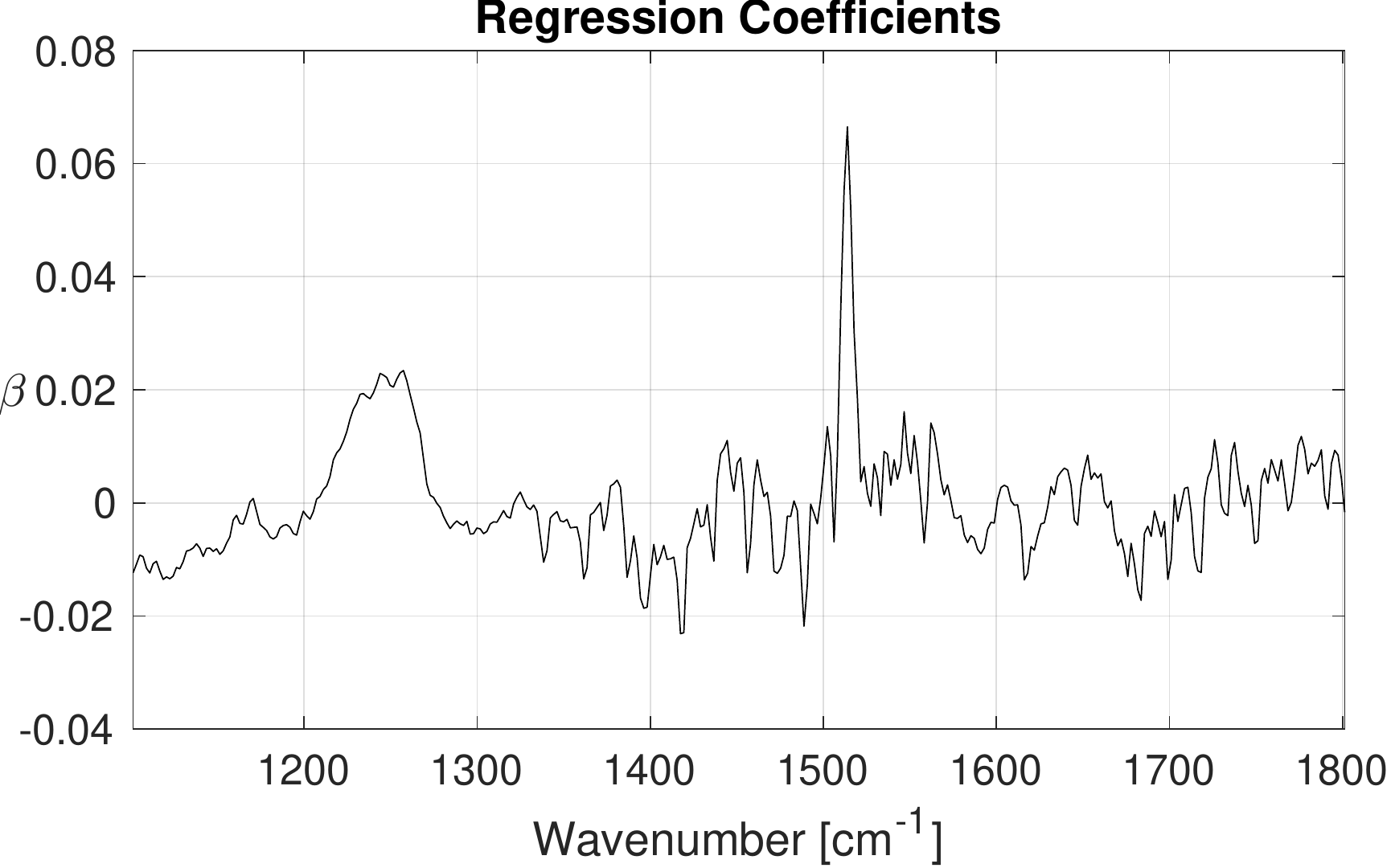}
    \begin{picture}(0,0)
        \put(20,0){\footnotesize{(c) Data split ignoring grouping}}
    \end{picture}
\end{minipage}
\caption{Three ways of splitting the ATR-FTIR spectral dataset (top) and the corresponding regression coefficients obtained by PLS (bottom), with number of components and RMSEs reported in Table \ref{tab:ilex_para}.}
\label{fig:Ill_Exp_Para}
\end{figure}
\noindent 
\begin{table}[H]
    \centering
    \begin{tabular}{|c|c|c|c|}\hline
         \begin{tabular}{c}Regression\\ Coefficients\end{tabular}& \begin{tabular}{c}Number of\\ Components\end{tabular} & RMSE Train & RMSE Test\\
        \hline \hline
        Fig.\ \ref{fig:Ill_Exp_Para}a & 7 & $1.4$$\times$$10^{-5}$ & $3.5$$\times$$10^{-5}$ \\
        \hline
        Fig.\ \ref{fig:Ill_Exp_Para}b & 7 & $1.4$$\times$$10^{-5}$ & $8$$\times$$10^{-4}$ \\ 
        \hline
        Fig.\ \ref{fig:Ill_Exp_Para}c & 8 & $1.5$$\times$$10^{-5}$ & $1.7$$\times$$10^{-5}$ \\ \hline
    \end{tabular}
    \caption{PLS regression coefficients, numbers of components, and RMSE for the training and test ATR-FTIR spectral datasets.}
    \label{tab:ilex_para}
\end{table}
The results of PLS regression for the ATR-FTIR spectral dataset shown in Fig.\ \ref{fig:Ill_Exp_Para} and Table \ref{tab:ilex_para} highlight implications of different ways of splitting data into training and test datasets in the presence of groups. 
In Fig.\ \ref{fig:Ill_Exp_Para}ab, groups of spectra belonging to the same concentration are randomly assigned to either the training or test dataset. In all cases, the root-mean-squared errors for the test data are higher than for the training data, due to the biased spectra corresponding to $C=0.014\, \text{g/g}$ that was chosen to be in the test dataset for all three splits. Fig.\ \ref{fig:Ill_Exp_Para}b shows a split where the test dataset contains groups with concentrations outside the training data. The root-mean-squared errors of prediction of the training and test datasets differ by one order of magnitude and show clearly that the model fails to extrapolate to the concentrations in the test dataset. It is recommended to assign the groups with spectra corresponding to the lowest and highest concentration as part of the training dataset, as is the case in Fig.\ \ref{fig:Ill_Exp_Para}a, so that the model interpolates rather than extrapolates.

Fig.\ \ref{fig:Ill_Exp_Para}c shows prediction results with randomized splitting of the training and test datasets. This violation of the grouping structure when splitting the dataset leads to {\em what appears to be} more accurate predictions for the test dataset (Table \ref{tab:ilex_para}). It is important to take into account that the prediction error for the test dataset obtained in case c approximates the expected generalization error only for new data {\em that are within the groups that are already part of the training data}. The reason for this restriction is that the between-group variance is higher than the within-group variance. Such an estimated prediction error will under-predict the prediction error when the model is applied to data that are not within the groups in the original dataset. In practice, the model is intended for use at {\em any} paracetamol concentration within the range of concentration that was in the training data, and the groupings constraint must be respected to obtain a model accuracy from the test set that is a reasonably accurate approximation of the generalization error. 

\subsection{Raman Spectral Data}
\noindent
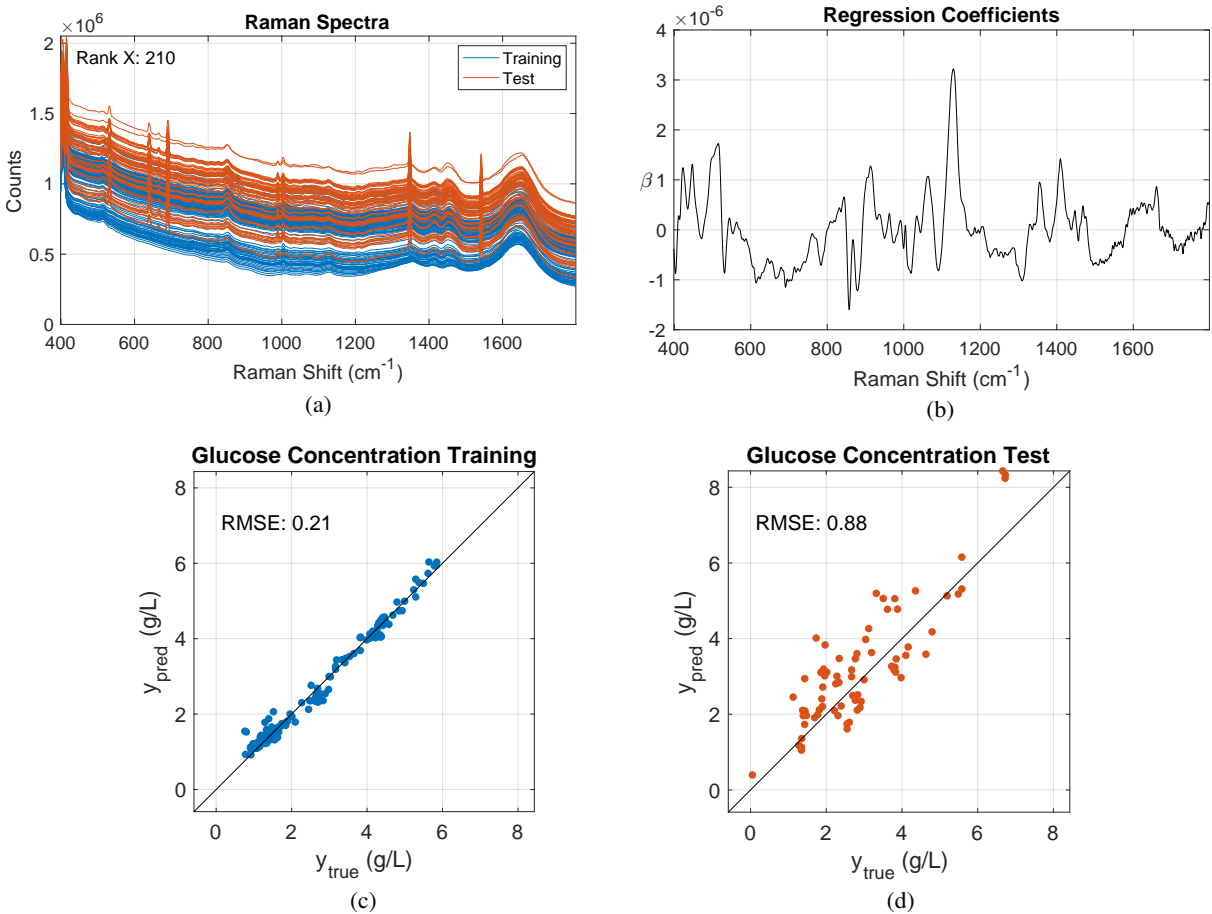
\begin{figure}[H]
\begin{center}
\begin{minipage}{0.46\textwidth}
    \includegraphics[width=\textwidth]{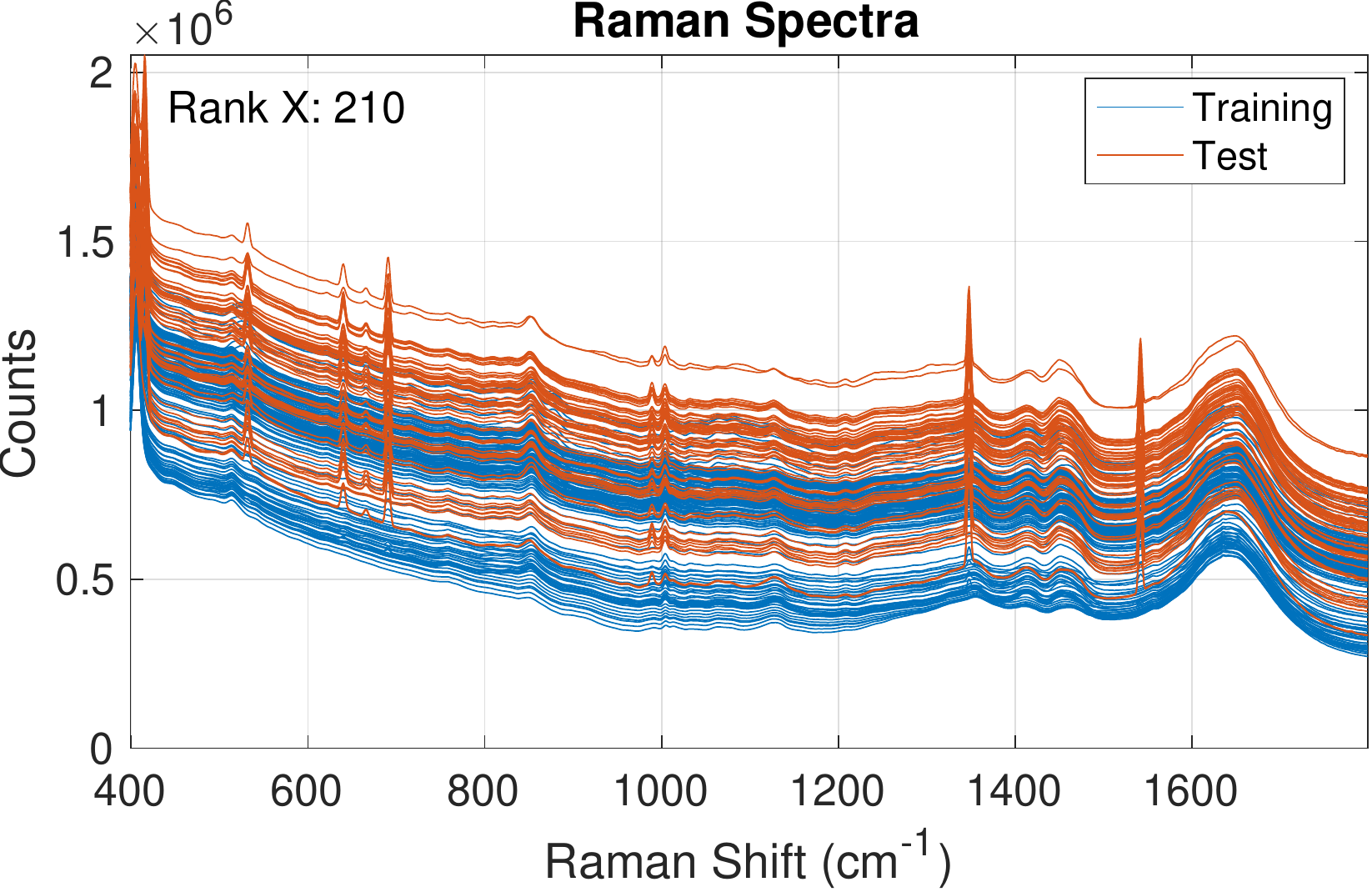}
\end{minipage}
\hspace{0.04\textwidth}
\begin{minipage}{0.46\textwidth}
    \includegraphics[width=\textwidth]{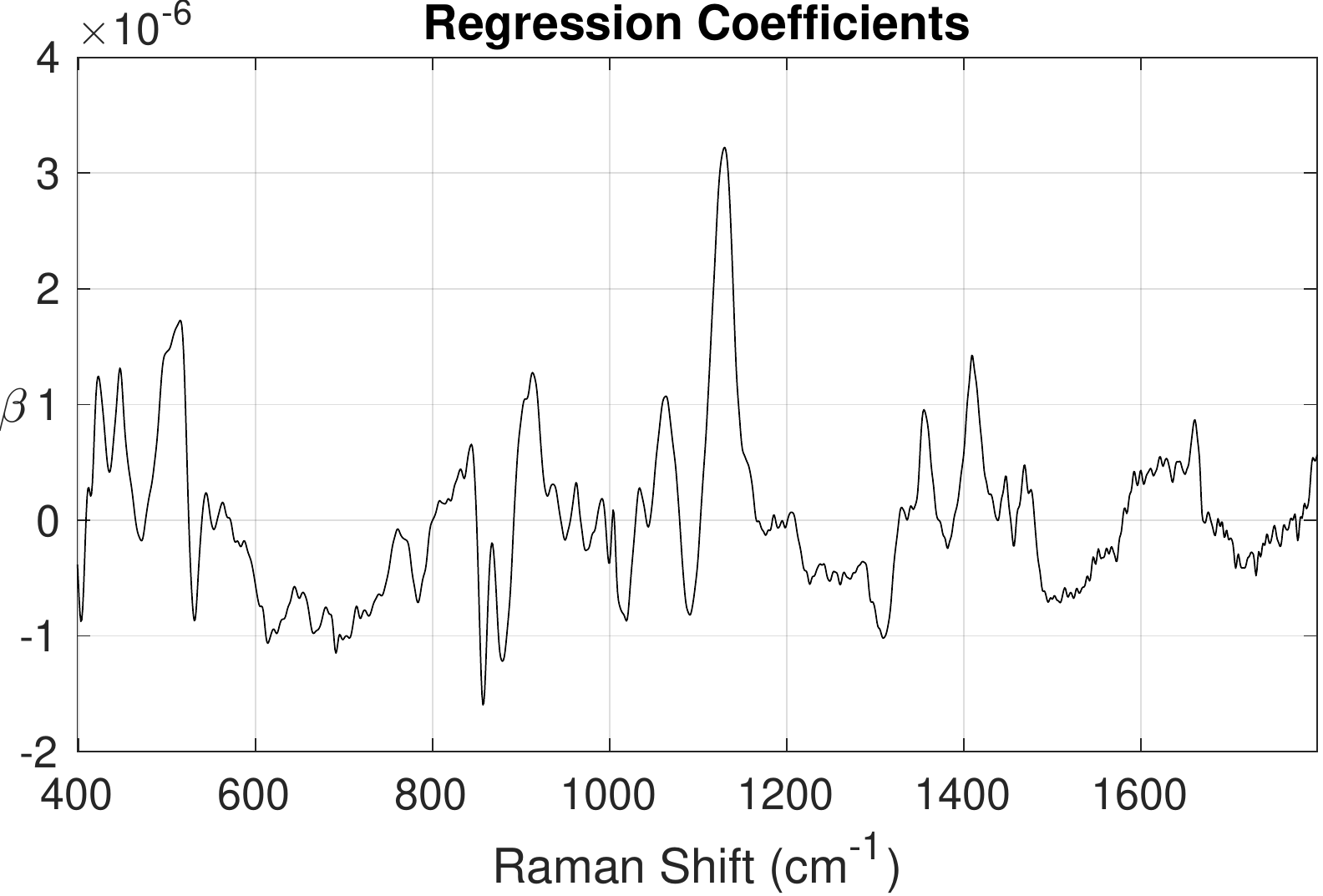}
\end{minipage}
\end{center}
\vspace{0.02\textwidth}
\begin{center}
    \begin{minipage}{0.32\textwidth}
        \includegraphics[width=\textwidth]{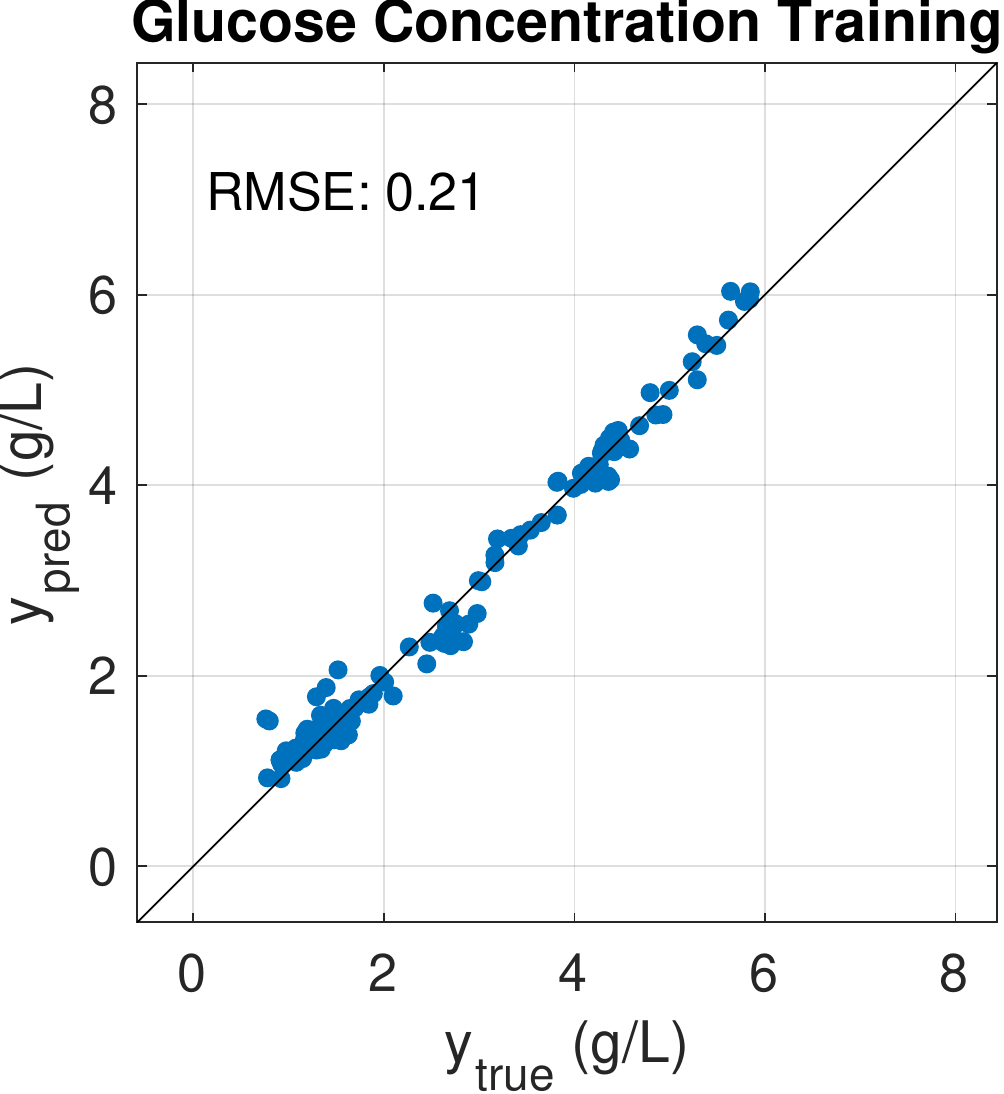}
    \end{minipage}
    \hspace{0.1\textwidth}
    \begin{minipage}{0.32\textwidth}
        \includegraphics[width=\textwidth]{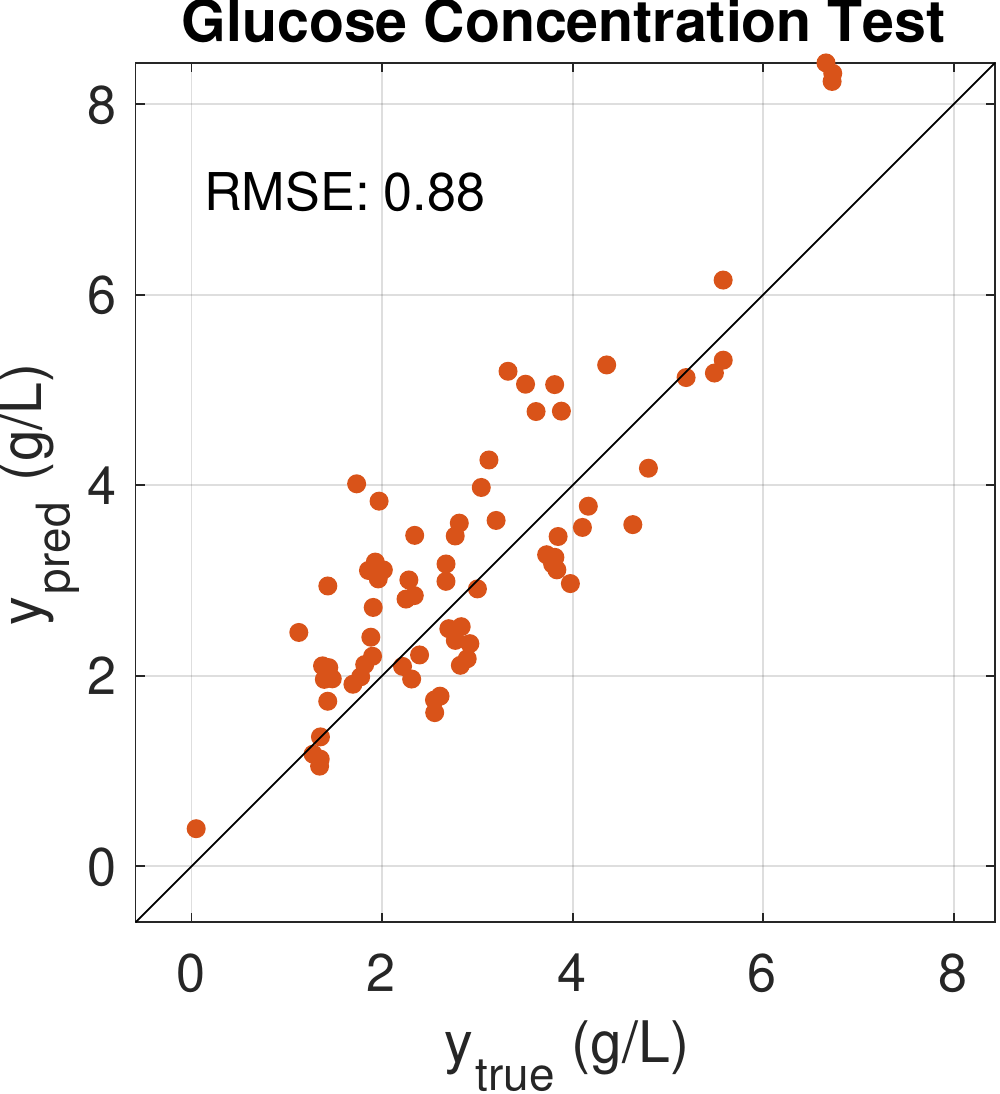}
    \end{minipage}
    \vspace{0.01\textwidth}\\
\end{center}
\begin{picture}(0,0)
    \put(120,188){{\footnotesize {(a)}}}
    \put(355,186){{\footnotesize {(b)}}}
    \put(137,0){{\footnotesize {(c)}}}
    \put(340,0){{\footnotesize {(d)}}}
\end{picture}
\caption{(a) Raman spectral dataset without background subtraction with the training data (bioreactor run 1) in blue and the test data (bioreactor run 2) in red, (b) the regression coefficients for a \acrshort{pls} model with 7 components, (c,d) the regression results for the training and test data.}
\label{fig:Ill_ex_ramangroup1}
\end{figure}

Figure \ref{fig:Ill_ex_ramangroup1} shows the results for a PLS model trained on data from bioreactor run 1, with the test set being data from bioreactor run 2 (Fig.\  \ref{fig:Ill_ex_ramangroup1}d). The regression coefficients show that the Raman shifts in the region around 1000--1200 cm$^{-1}$ are most important for predicting glucose concentration. The model predicts the glucose concentration in the test dataset fairly well, however, the model accuracy is significantly higher on the training dataset than on the test dataset. 

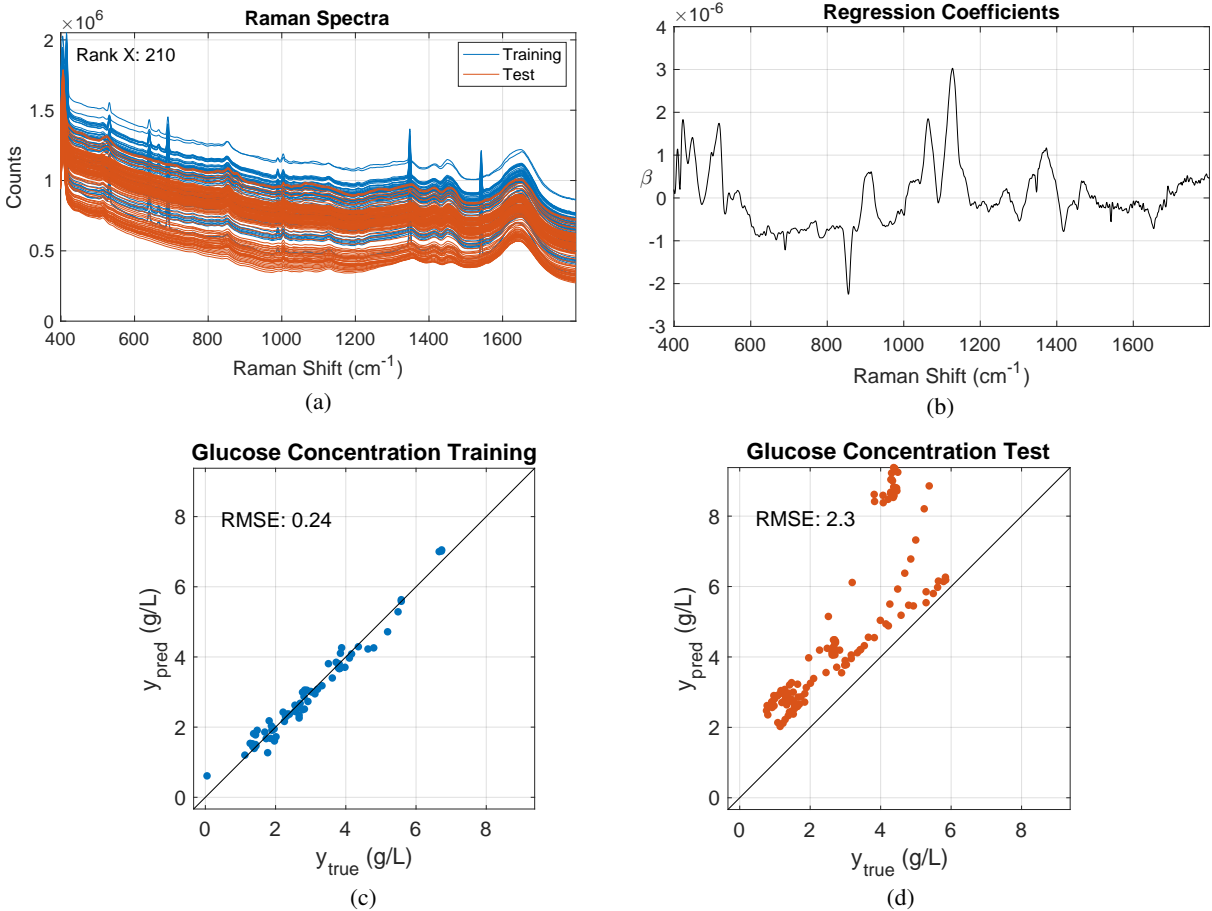
\begin{figure}[!htb]
\begin{center}
\begin{minipage}{0.46\textwidth}
    \includegraphics[width=\textwidth]{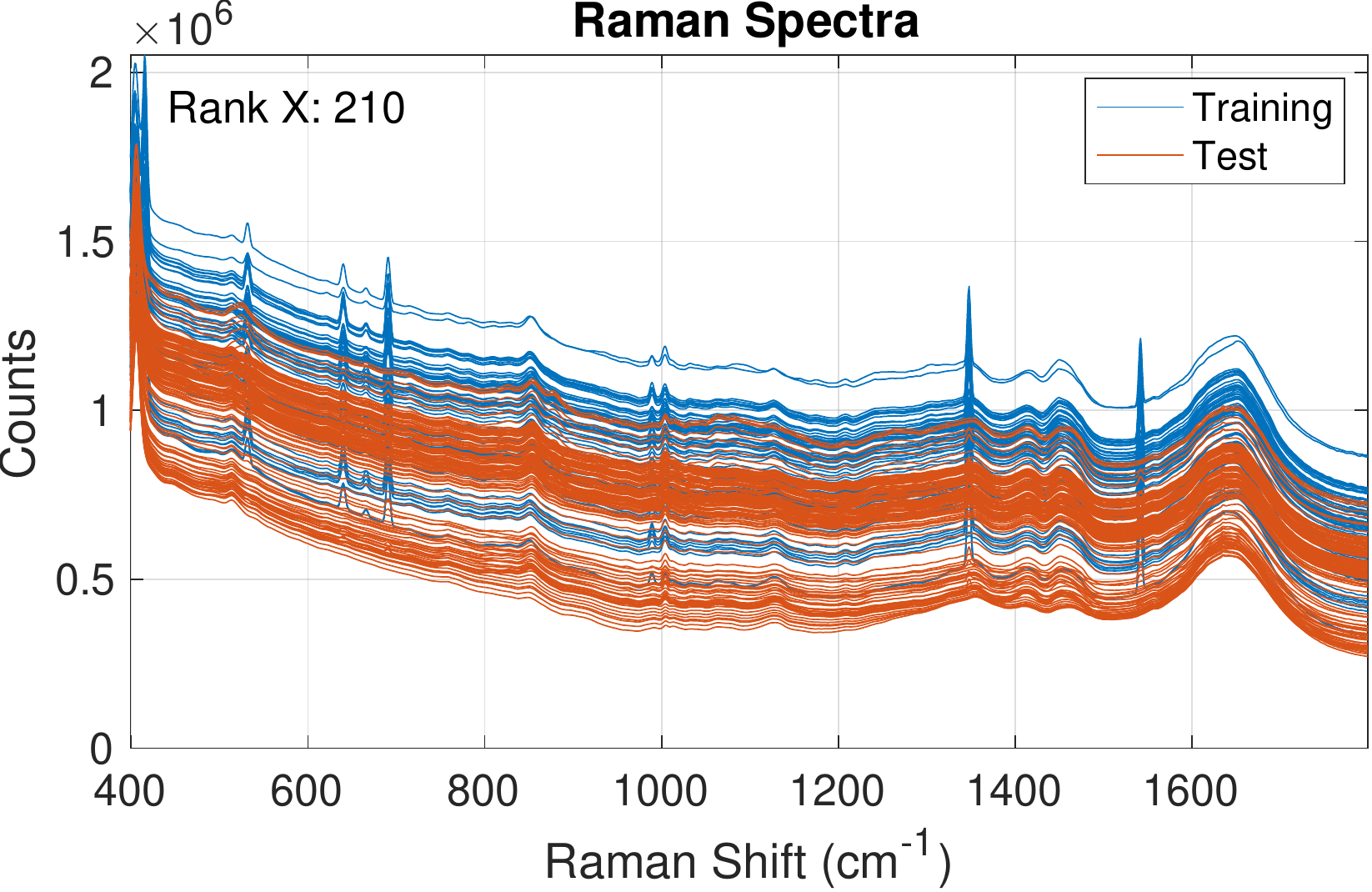}
\end{minipage}
\hspace{0.04\textwidth}
\begin{minipage}{0.46\textwidth}
    \includegraphics[width=\textwidth]{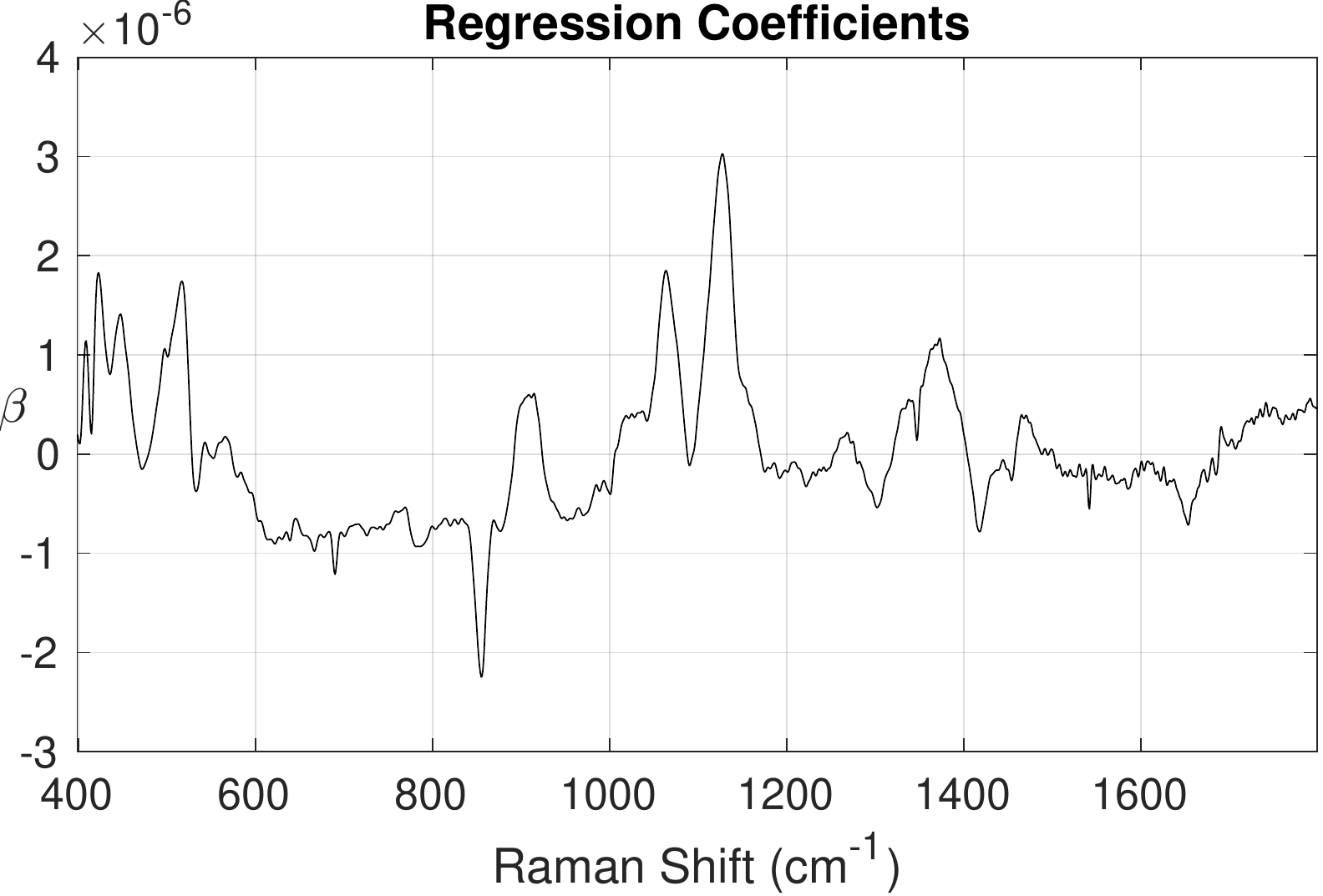}
\end{minipage}
\end{center}
\vspace{0.02\textwidth}
\begin{center}
    \begin{minipage}{0.32\textwidth}
        \includegraphics[width=\textwidth]{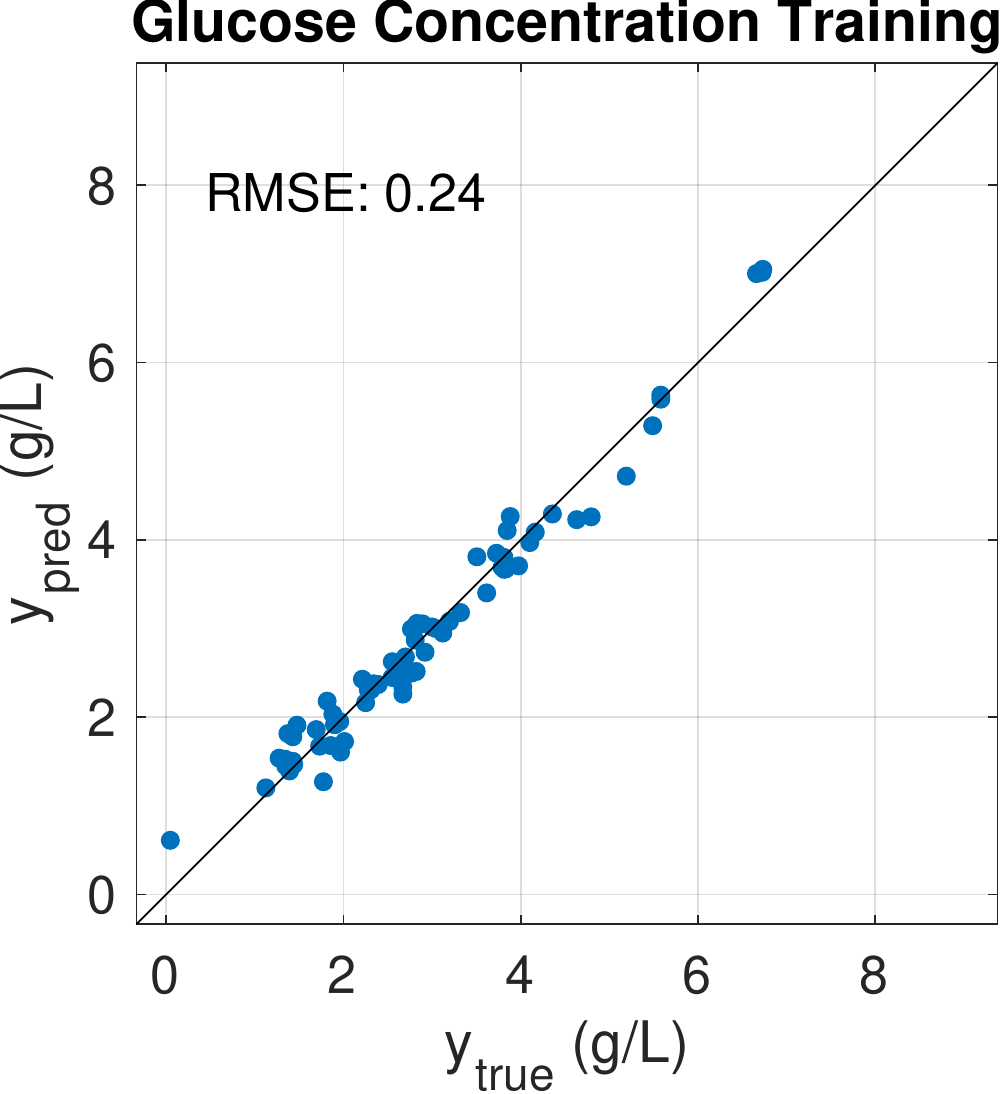}
    \end{minipage}
    \hspace{0.1\textwidth}
    \begin{minipage}{0.32\textwidth}
        \includegraphics[width=\textwidth]{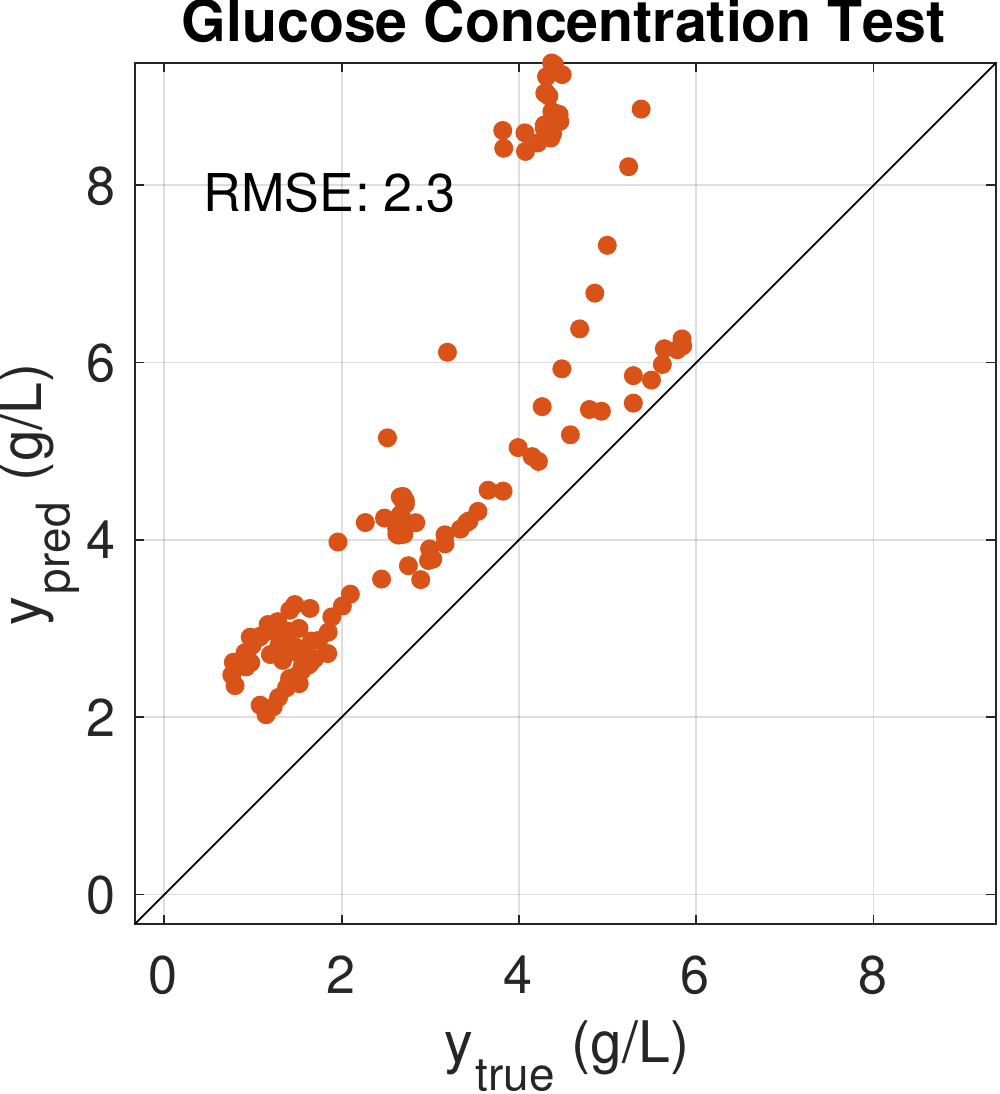}
    \end{minipage}
    \vspace{0.01\textwidth}\\
\end{center}
\begin{picture}(0,0)
    \put(120,188){{\footnotesize {(a)}}}
    \put(355,186){{\footnotesize {(b)}}}
    \put(137,0){{\footnotesize {(c)}}}
    \put(340,0){{\footnotesize {(d)}}}
\end{picture}
\caption{(a) Raman spectral dataset without background subtraction with the training data (bioreactor run 2) in blue and the test data (bioreactor run 1) in red, (b) the regression coefficients for a \acrshort{pls} model with 6 components, (c,d) the regression results for the training and test data.}
\label{fig:Ill_ex_ramangroup2}
\end{figure}

Figure \ref{fig:Ill_ex_ramangroup1} shows the results for a PLS model trained on data from bioreactor run 2, with the test set being data from bioreactor run 1 (Fig.\  \ref{fig:Ill_ex_ramangroup2}d). Most of the glucose concentrations predicted by the model are biased and the prediction errors are much larger than estimated by the model that was trained on bioreactor run 1 and predicted on bioreactor run 2 (cf.\ Fig.\ \ref{fig:Ill_ex_ramangroup1}d). However, the regression coefficients determined from bioreactor run 2 are very similar to the regression coefficients determined from bioreactor run 1. Ensuring a strict split of the data, in which data from each bioreactor are only in the training or test datasets, is advised to obtain a more realistic generalization error of the model. As seen in the ATR-FTIR spectral dataset example, grouping should be respected when building such models, due to the between-group variation being higher than the within-group variation. The large test errors in Fig.\ \ref{fig:Ill_ex_ramangroup2} illustrate the challenges that occur when trying to build models that will predict glucose concentration well for new bioreactor runs. The intrinsically very low photon efficiency of Raman spectroscopy and batch-to-batch and time-varying concentrations of other species makes it challenging to build highly accurate predictive models of glucose in bioreactors.

\begin{figure}[!htb]
\begin{center}
\begin{minipage}{0.46\textwidth}
    \includegraphics[width=\textwidth]{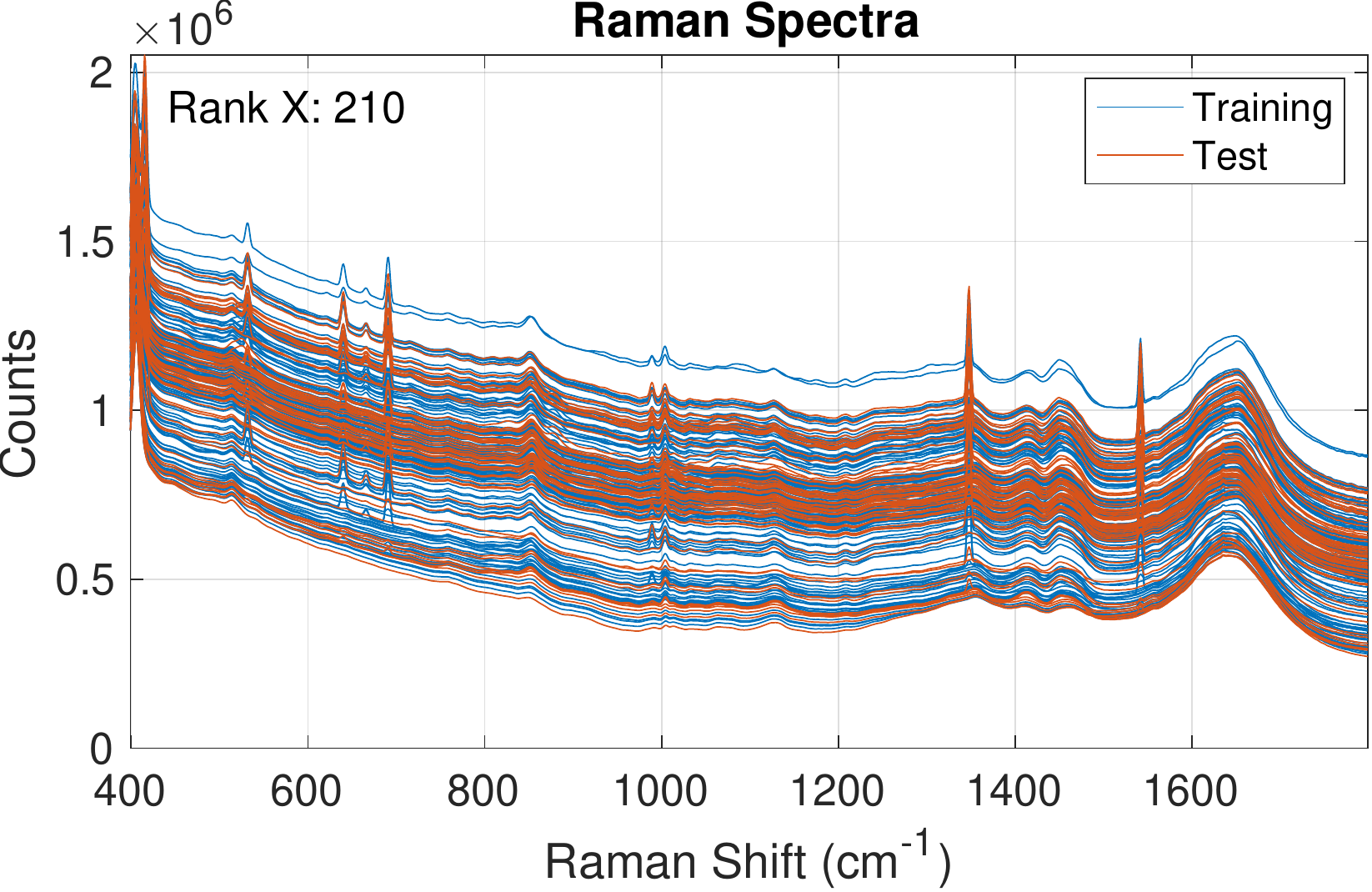}
\end{minipage}
\hspace{0.04\textwidth}
\begin{minipage}{0.46\textwidth}
    \includegraphics[width=\textwidth]{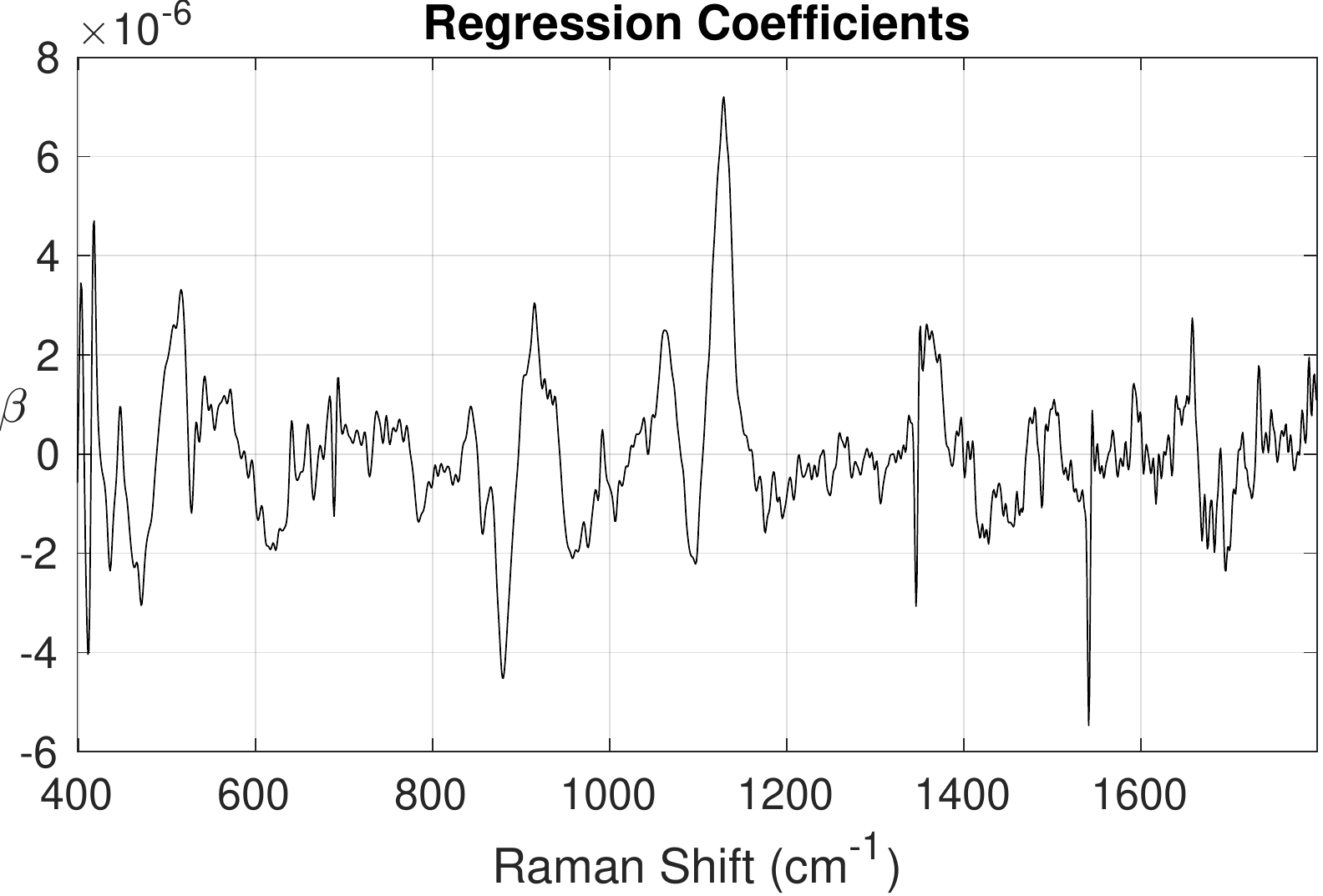}
\end{minipage}
\end{center}
\vspace{0.02\textwidth}
\begin{center}
    \begin{minipage}{0.32\textwidth}
        \includegraphics[width=\textwidth]{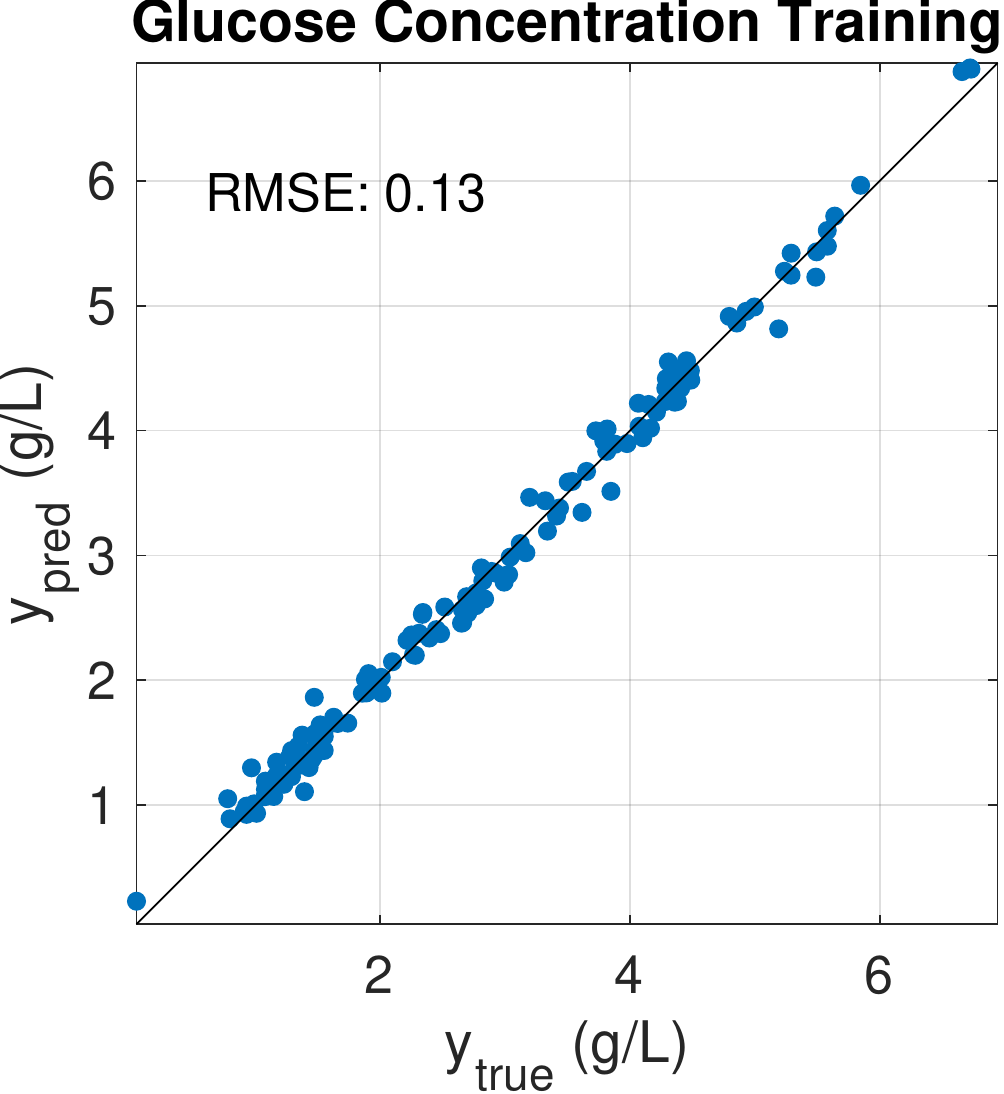}
    \end{minipage}
    \hspace{0.1\textwidth}
    \begin{minipage}{0.32\textwidth}
        \includegraphics[width=\textwidth]{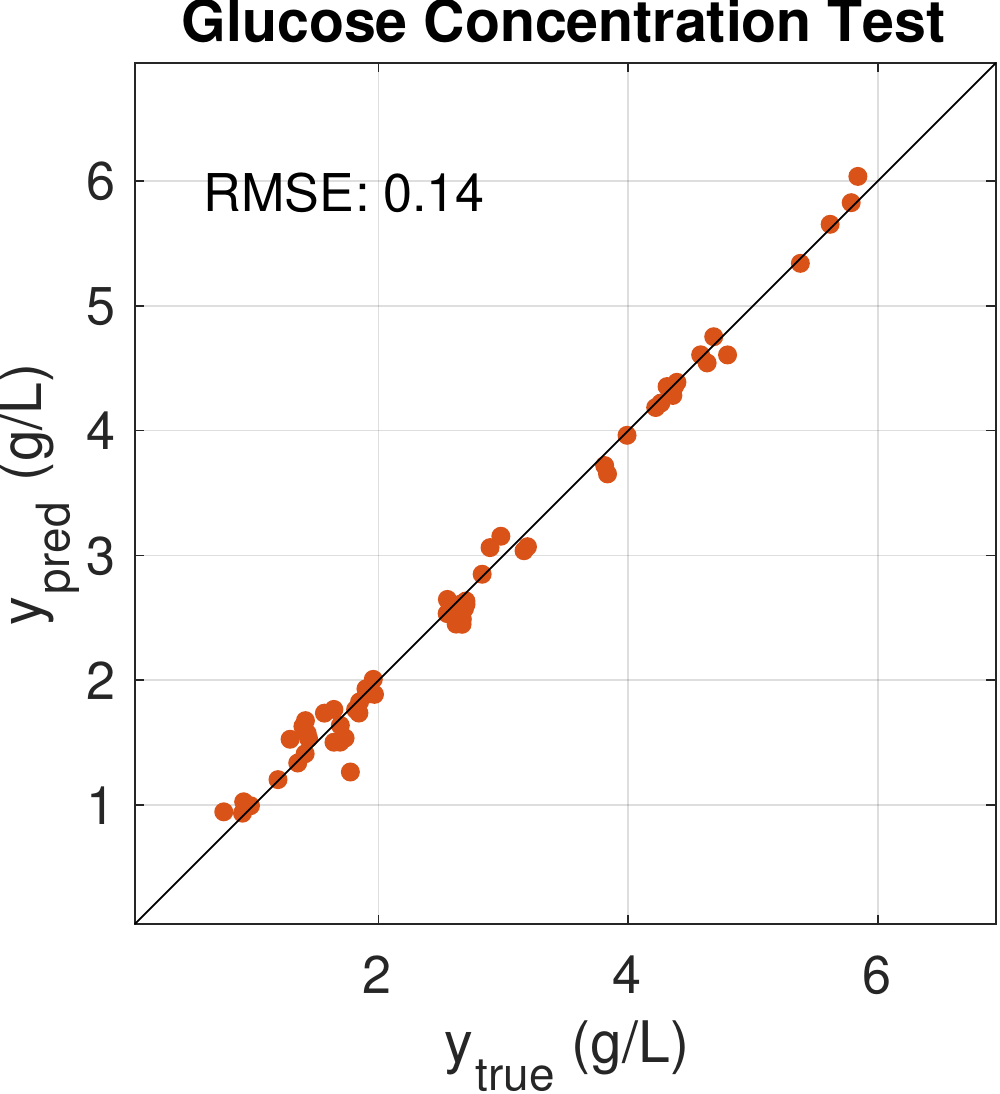}
    \end{minipage}
    \vspace{0.01\textwidth}\\
\end{center}
\begin{picture}(0,0)
    \put(120,188){{\footnotesize {(a)}}}
    \put(355,186){{\footnotesize {(b)}}}
    \put(137,0){{\footnotesize {(c)}}}
    \put(340,0){{\footnotesize {(d)}}}
\end{picture}
\caption{(a) Raman spectral dataset without background subtraction with the training data in blue and the test data in red using a random split, (b) the regression coefficients for a \acrshort{pls} model with 12 components, (c,d) the regression results for the training and test data.}
\label{fig:Ill_ex_ramannogroup}
\end{figure}
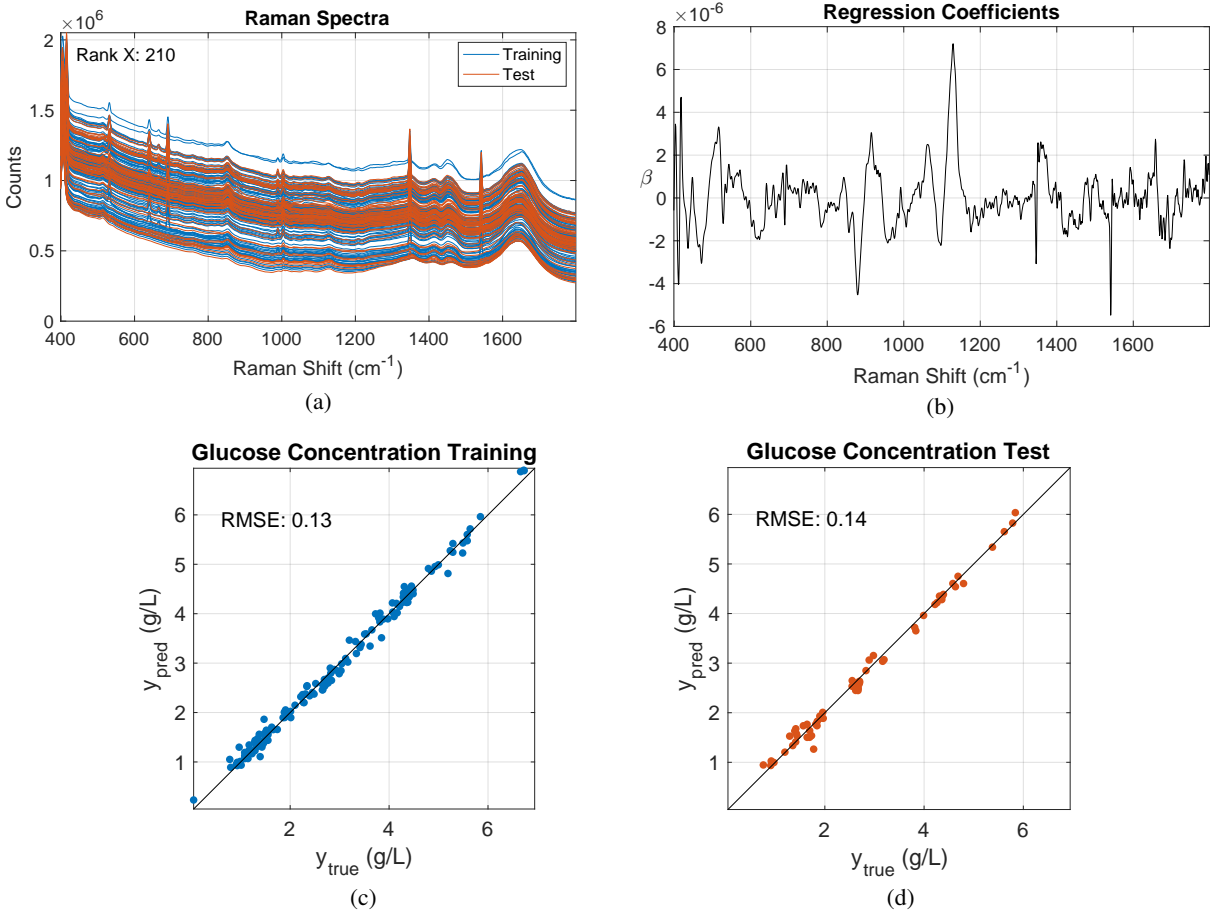

Figure \ref{fig:Ill_ex_ramannogroup} shows the results of building a PLS model for Raman spectra from multiple bioreactor runs. The data are split randomly, leading to data from all of the bioreactors being present in both the training and test datasets. The model accurately predicts the glucose concentration in the test dataset. The number of PLS components is significantly higher than in the cases of Fig.\ \ref{fig:Ill_ex_ramangroup1} and \ref{fig:Ill_ex_ramangroup2}, but overall the regression coefficients are similar. However the regression coefficients in Fig.\ \ref{fig:Ill_ex_ramannogroup}) have a higher level of noise, suggesting that the model overfits to specific conditions being present in the respective bioreactors, i.e., that too many components were used.
The split leads to {\em what appears to be} more accurate predictions for the test dataset similar to the ATR-FTIR spectral dataset. It is important to take into account that the prediction error for the test dataset obtained approximates the expected generalization error only for new data {\em that are within the reactor runs that are already part of the training data}. In addition, the random split also does not respect the time series structure of the dataset and is thus not a fair estimate of the model accuracy for reactor conditions further in the future (i.e., not in between conditions present in the training dataset as it is the case here for most data in the test dataset). 

\FloatBarrier
\subsection{LFP Battery Dataset} 
\noindent
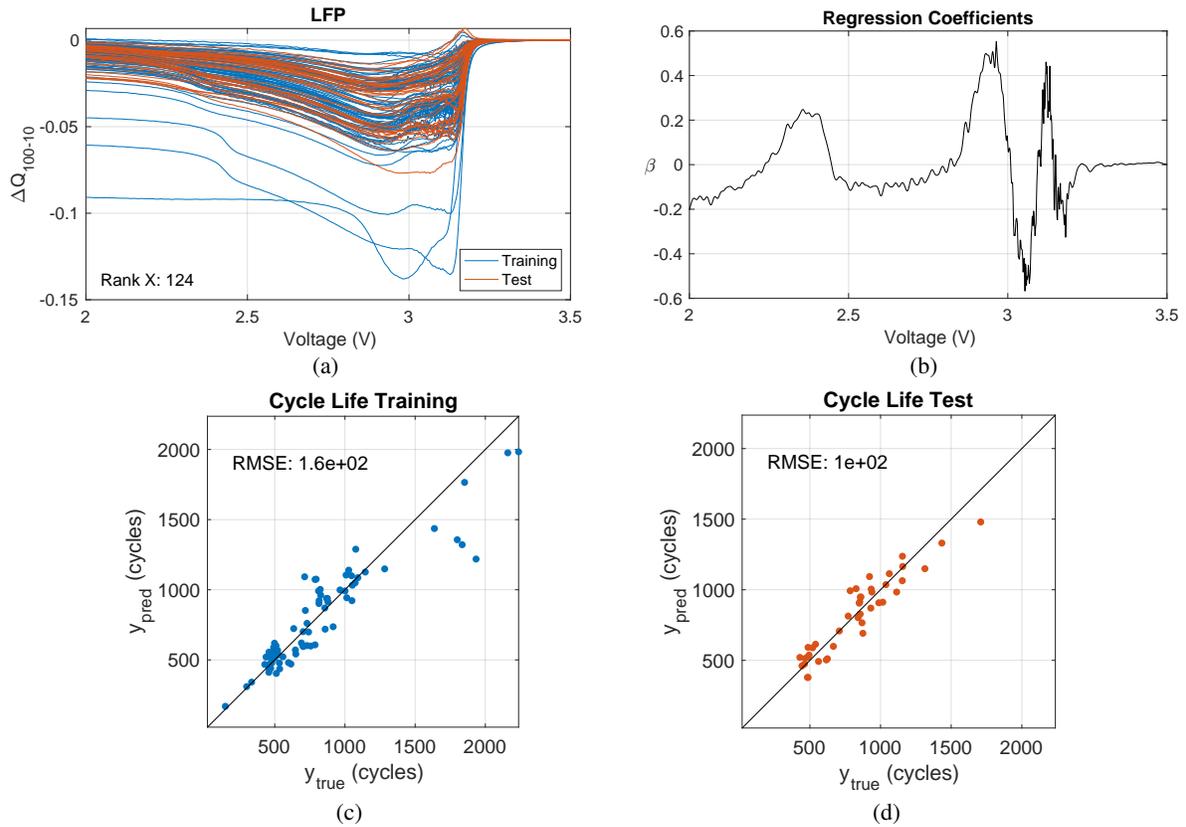
\begin{figure}[!htb]
\begin{center}
\begin{minipage}{0.46\textwidth}
    \includegraphics[width=\textwidth]{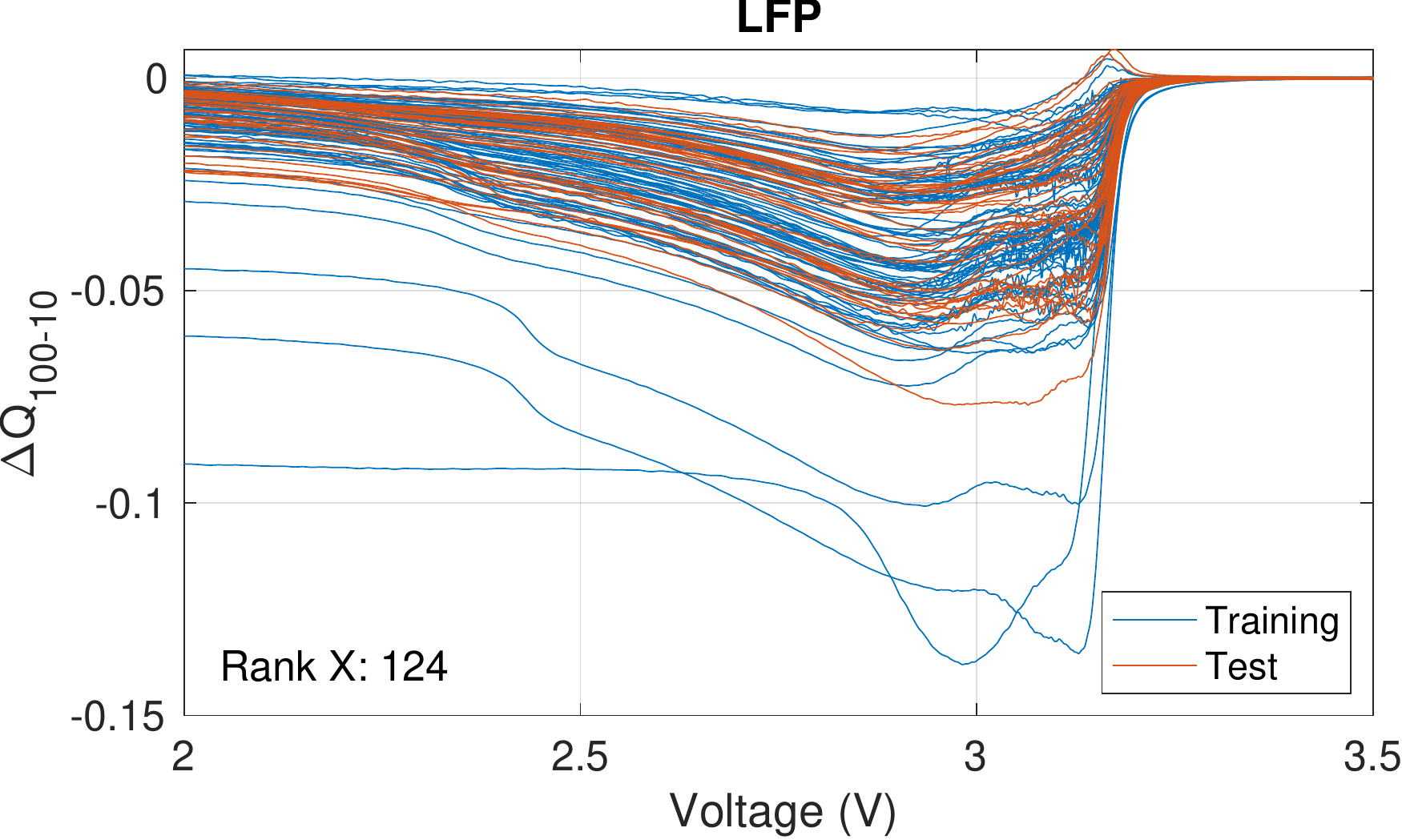}
\end{minipage}
\hspace{0.04\textwidth}
\begin{minipage}{0.43\textwidth}
    \includegraphics[width=\textwidth]{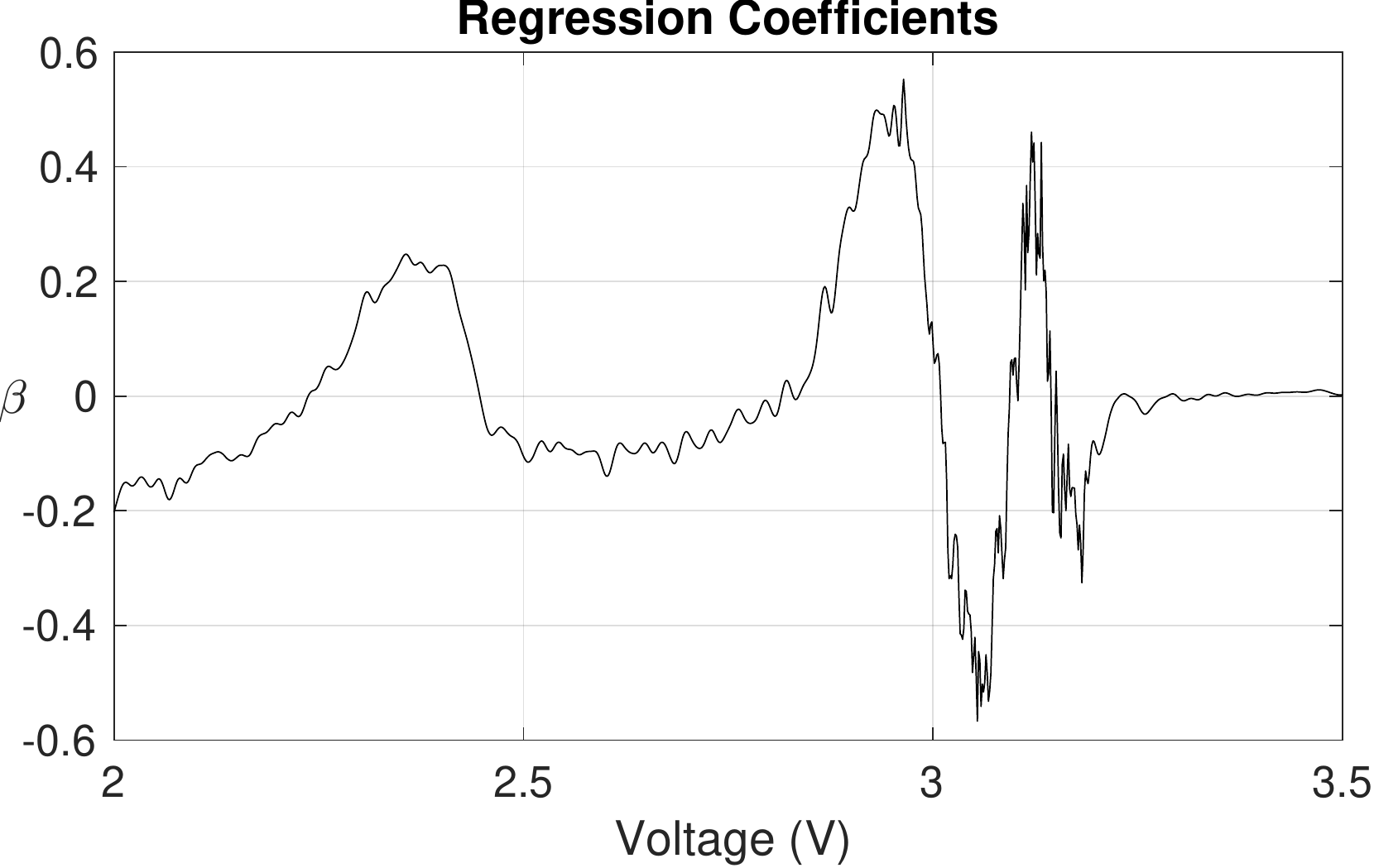}
\end{minipage}
\end{center}
\vspace{0.01\textwidth}
\begin{center}
    \begin{minipage}{0.32\textwidth}
        \includegraphics[width=\textwidth]{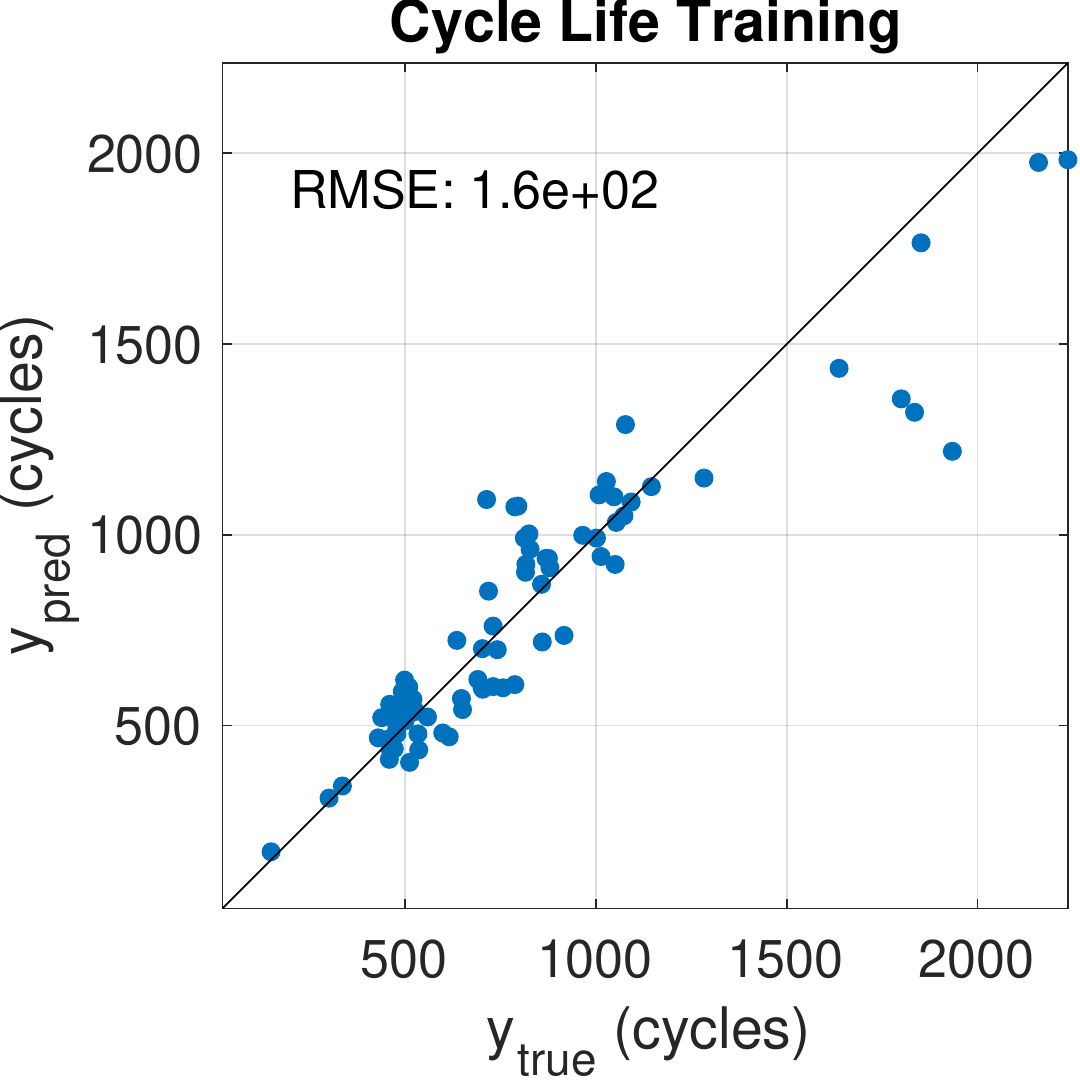}
    \end{minipage}
    \hspace{0.1\textwidth}
    \begin{minipage}{0.32\textwidth}
        \includegraphics[width=\textwidth]{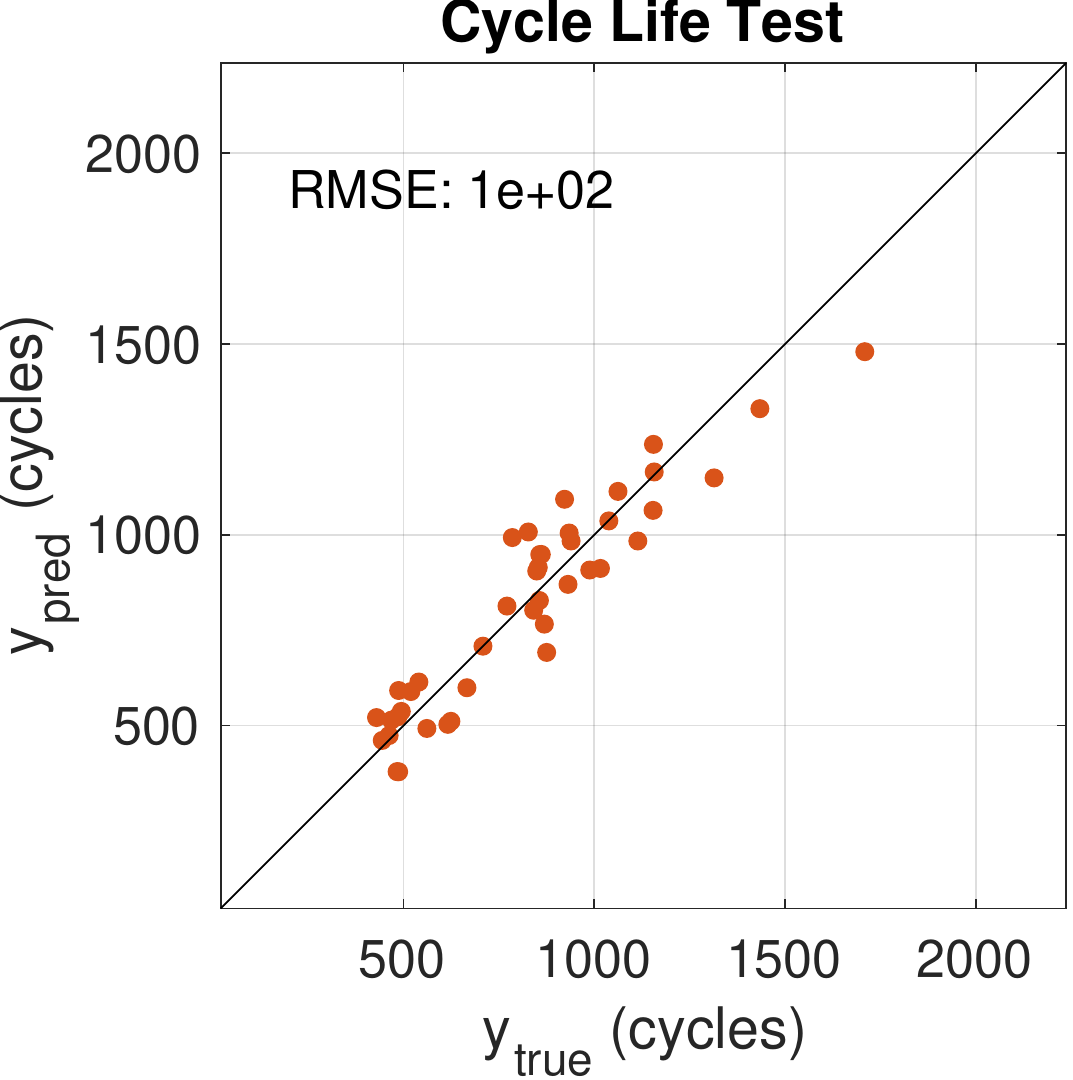}
    \end{minipage}
    \vspace{0.01\textwidth}\\
\end{center}
\begin{picture}(0,0)
    \put(128,169){{\footnotesize {(a)}}}
    \put(354,169){{\footnotesize {(b)}}}
    \put(137,0){{\footnotesize {(c)}}}
    \put(340,0){{\footnotesize {(d)}}}
\end{picture}
\caption{(a) \acrshort{lfp} dataset with the training data in blue and the test data in red, (b) the regression coefficients for a \acrshort{pls} model with four components, (c,d) the regression results for the training and test data.}
\label{fig:Ill_ex_lfp}
\end{figure}
The distribution of cycle life of batteries in the \acrshort{lfp} dataset is non-Gaussian, with a high amount of cells with medium cycle life, three cells with a significantly shorter cycle life, and a few cells with significantly longer cycle life. This distribution results in the root-mean-squared-errors varying depending on precisely which data are in the training and test datasets, even with well-tuned hyperparamaters. For the split in Fig.\ \ref{fig:Ill_ex_lfp}, the \Acrshort{rmse} for the test data is lower than for the training data, which is because the test dataset contains only medium- to short-lived cells that are well captured by the model. The RMSE for the training dataset for medium- to short-lived cells in Fig.\ \ref{fig:Ill_ex_lfp} is similar to the RMSE for the test dataset. 
The RMSE for the training data is strongly influenced by the higher \acrshort{rmse} of the longest lived cells. This example illustrates the significant effect that a small number of data points at extreme values of the data distribution, although reasonably consistent with the rest of the data, can have on the estimated prediction errors.
\begin{figure}[!htb]
\centering
    \begin{tikzpicture}
        \node[above right, inner sep=0](image) at (0,0) {
            \includegraphics[width=0.5\textwidth]{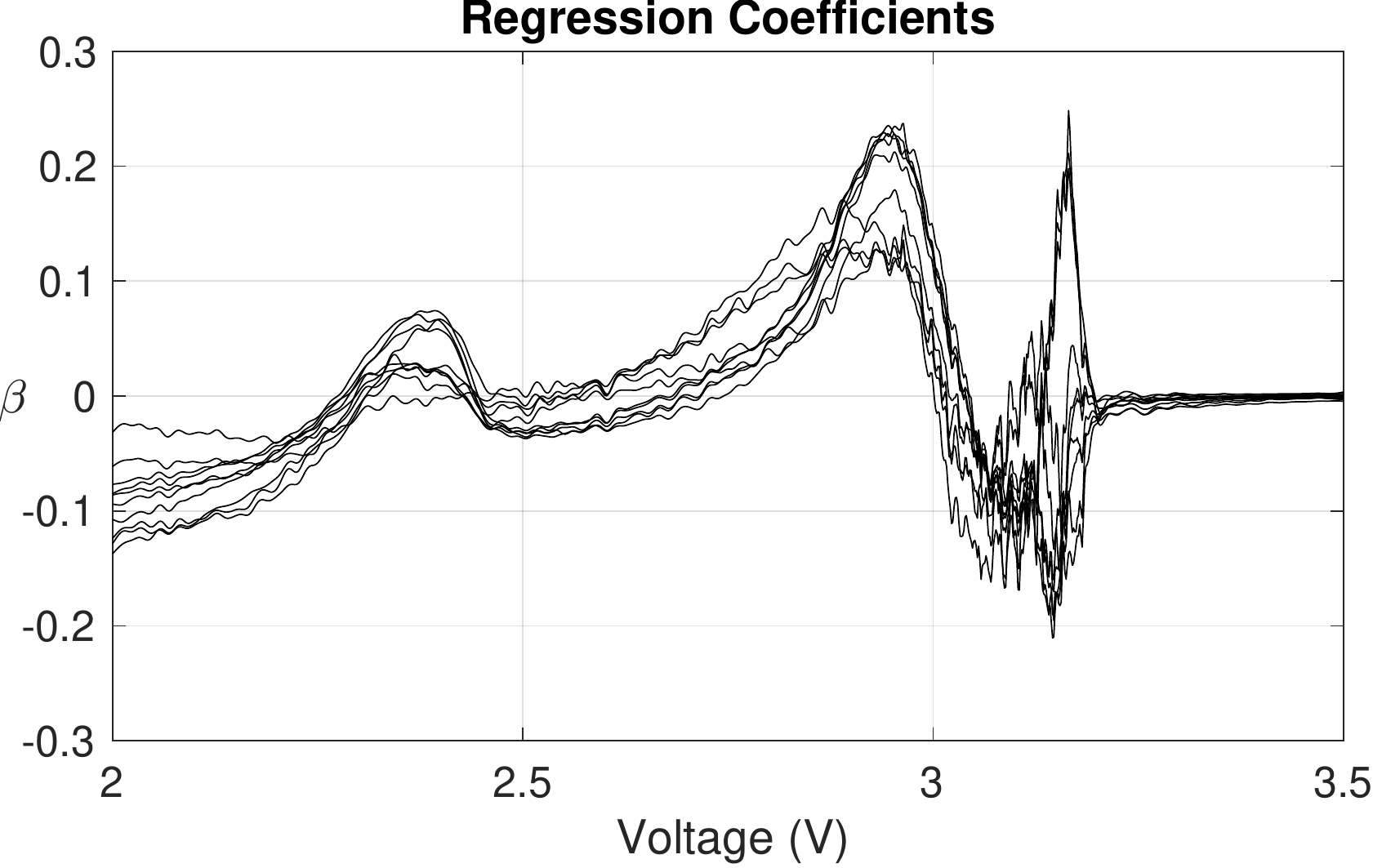}
        };
        \begin{scope}[
        x={($0.1*(image.south east)$)},
        y={($0.1*(image.north west)$)}]
        \ifnum \value{develop}>0
        	\draw[lightgray,step=0.5] (image.south west) grid (image.north east);
        
        	\foreach \x in {0,1,...,10} { \node [below] at (\x,0) {\x}; }
        	\foreach \y in {0,1,...,10} { \node [left] at (0,\y) {\y};}
        \fi
        \draw[thick,red] (6.75,1.5) rectangle (8.25,9.35);
        \end{scope}
    \end{tikzpicture}
\caption{Varying regression coefficients for different train-test splits.}
\label{fig:var_reg_coef}
\end{figure}
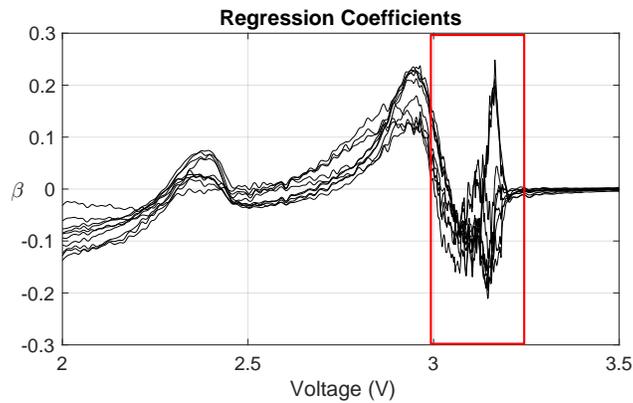

Figure \ref{fig:var_reg_coef} shows the regression coefficients for models that were trained on ten training-test splits of the dataset. The resulting coefficients vary significantly, particularly in the region around 3 volts marked by a red box. This high variability suggests that PLS finds it challenging to extract the information contained in this voltage region. Reasons for this behaviour could be nonlinear effects or inherent biases due to battery physics that manifest in this voltage region. For more insights, we refer to the supplementary information of \cite{attia2020closed} that contains more details about the measuring campaign and corresponding differences in calendar aging due to different storage times of batteries before the start of the measurements.

%% file: 6Conclusion.tex
\clearpage
\section{Conclusion}
\label{sec:Conclusion}
The \acrshort{lavade} software enables users to easily compare different latent variable regression and alternative regression methods when applied to four datasets of different complexity ranging from a synthetic dataset to real-world datasets. 
The visualizations of the data, train-test splits, regression coefficients, and predictions are useful to understanding in these regression methods. These visualizations reveal important details about the data and implications of the model building strategy. The interactive environment of the \acrshort{gui} fosters a deeper understanding of how models interact with data. Comparing regularization parameters of \acrshort{lasso}, \acrshort{rr}, \acrshort{en}, and the number of components for \acrshort{pls} and \acrshort{pcr} helps to understand how the different approaches of regularization affect the resulting regression coefficients. 

The open-source \acrshort{lavade} software intends to be a starting point to gain intuition on latent variable and other multivariate regression methods. Ultimately we hope to inspire teachers to use this tool to advance their teaching methods and encourage extensions of the software and examples.
    


%% file: Aknowledgements.tex
\section*{Code and Data Availability}
The code is available at \small\url{https://github.com/JoachimSchaeffer/LAVADE}. The repository contains the datasets that are described in this article. The paracetamol dataset was recorded and an analysis of the dataset was published with \cite{fujiwara2002paracetamol}. The repository contains only a small subset of the \acrshort{lfp} dataset that was published with \cite{severson2019data}.
The entire \acrshort{lfp} dataset is available at \small\url{https://data.matr.io/1/}.

\section*{Acknowledgments}
Financial support is acknowledged from the German Academic Exchange Service (DAAD) for Joachim Schaeffer and from the U.S. Food and Drug Administration, Grant \#U01FD006483, for Richard Braatz. Eric Lenz from the Control and Cyber-Physical Systems Laboratory at TU Darmstadt is thanked for providing valuable feedback and reviewing the paper before submission. Alexis Dubs and Nili Persits at MIT are thanked for providing information on the Raman spectral data and the associated experimental data.